\documentclass[a4paper]{cas-sc}

\usepackage[scaled]{stmaryrd}
\usepackage{indentfirst}
\usepackage{booktabs}
\usepackage{tabularx}
\usepackage{graphicx}
\usepackage{adjustbox}
\usepackage{amsmath}
\usepackage{array}
\usepackage{algorithm}
\usepackage{algpseudocode}
\usepackage{float}

\usepackage{marvosym}
\usepackage{xcolor}
\usepackage{amsthm}    
\usepackage{longtable} 
\usepackage{natbib}
\usepackage{multirow}

\ExplSyntaxOn
\cs_set:Npn \__first_footerline: {}
\ExplSyntaxOff

\def\tsc#1{\csdef{#1}{\textsc{\lowercase{#1}}\xspace}}
\tsc{WGM}
\tsc{QE}
\tsc{EP}
\tsc{PMS}
\tsc{BEC}
\tsc{DE} 

\begin{document}
\let\WriteBookmarks\relax
\def\floatpagepagefraction{1}
\def\textpagefraction{.001}
\let\printorcid\relax
\shorttitle{Enhancing Mixed Oil Length Prediction through Graph Representation Learning}
\shortauthors{Xinyi Hong et~al.} 
\title [mode = title]{Enhancing Mixed Oil Length Prediction through Graph Representation Learning}

\author[ecnu,sjtu]{Xinyi Hong}
\author[nyu]{Hai Shu}
\author[ecnu]{Zhaoyang Zhang}
\author[cupec]{Lei Chen}
\author[ecnu,ism,sokendai]{Yanfeng Yang}\cormark[1]\ead{yanfengyang0316@gmail.com}
\author[ecnu]{Ziqi Chen}\cormark[1]\ead{zqchen@fem.ecnu.edu.cn}
\cortext[1]{Corresponding Author}

\affiliation[ecnu]{organization={East China Normal University}, addressline={No. 3663, North Zhongshan Road}, city={Shanghai}, postcode={200062}, country={China}}
\affiliation[sjtu]{
    organization={Shanghai Jiao Tong University},
    addressline={No. 800 Dongchuan Road, Minhang District},
    city={Shanghai},
    postcode={200240},
    country={China}
}
\affiliation[nyu]{organization={New York University}, addressline={708 Broadway}, city={New York}, postcode={10003}, country={United States of America}}
\affiliation[cupec]{organization={China University of Petroleum (East China)}, addressline={No.66, West Changjiang Road, Huangdao District}, city={Qingdao}, postcode={266580}, country={China}}
\affiliation[ism]{organization={The Institute of Statistical Mathematics
},
            addressline={10-3 Midori-cho, Tachikawa
}, 
            city={Tokyo},
            postcode={190-8562}, 
            country={Japan}}

\affiliation[sokendai]{organization={The Graduate University for Advanced Studies, SOKENDAI},
            addressline={Shonan Village, Hayama}, 
            city={Kanagawa},
            postcode={240-0193}, 
            country={Japan}}

\begin{abstract}
During the pipeline transportation of refined oil products, the inevitable formation of mixed oil segments poses considerable economic challenges to pipeline networks. Therefore, it is important to accurately predict the length of these mixed oil segments. Although previous methods have achieved relatively accurate predictions by integrating machine and deep learning algorithms, they ignore the statistical dependencies that remain among real-world mixed oil samples, which degrades model accuracy. To address this issue, we propose a novel Sample Dependencies Incorporated Graph Representation Learning (SDI-GRL) model.  By representing samples and pipeline features as nodes of a graph, SDI-GRL connects dependent samples through their shared features, allowing message passing to propagate information across related samples. This graph-based framework allows our model to capture the statistical dependencies among samples absent in prior works. A case study on a mixed oil dataset demonstrates that our SDI-GRL model surpasses the strongest existing approach, reducing both the RMSE and the MAE by more than 10\%.
\end{abstract}


\maketitle
\label{title}

 \hypersetup{
     citecolor=red, 
 }
\section{Introduction}
\label{sec:intr}

\begin{figure}[pos=htb]
    \centering
    \includegraphics[width=1\textwidth,trim=0 50 0 100,clip]{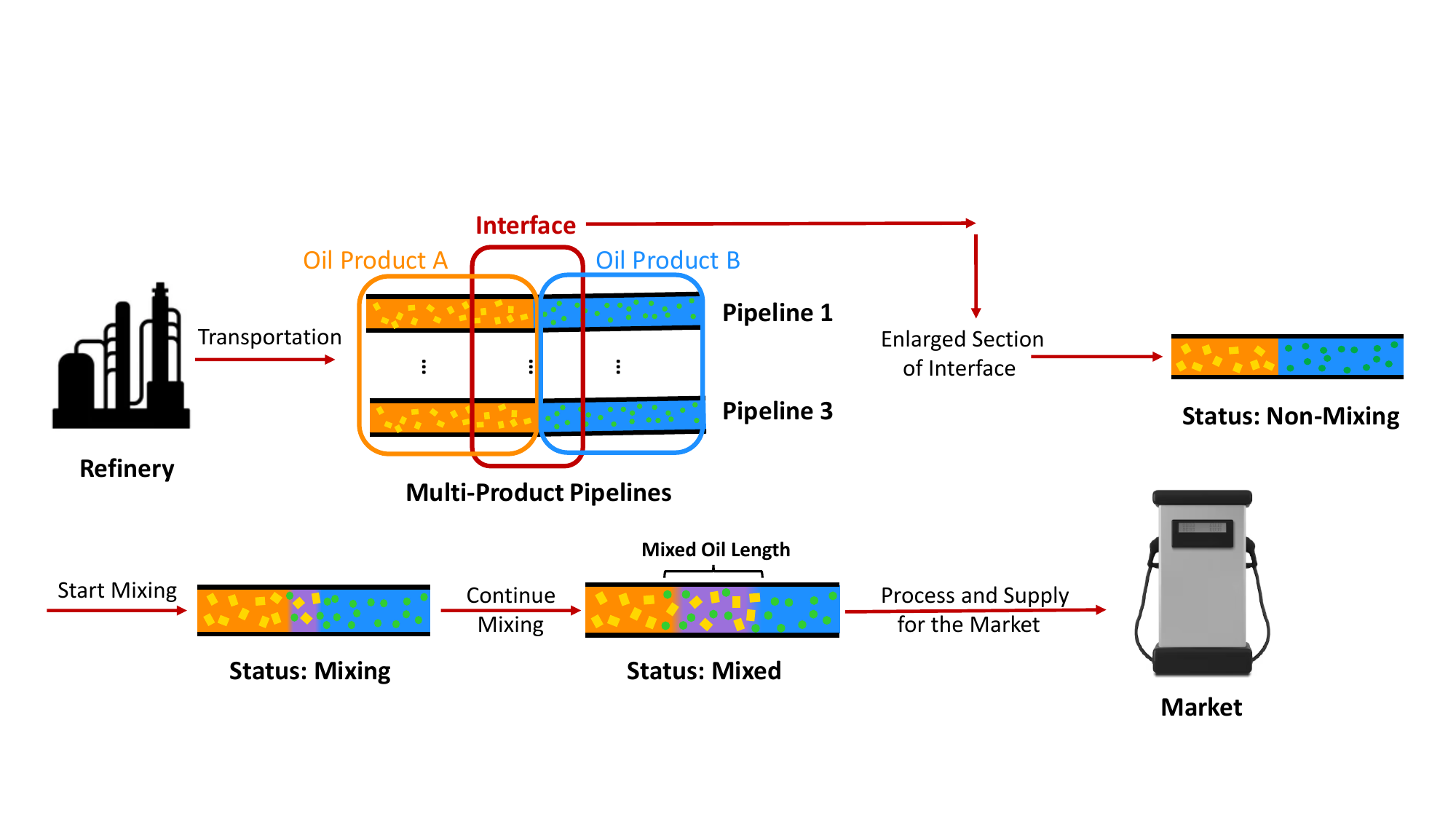}
    \caption{
        \textbf{Mixed Oil Formation Phenomenon During Transport.} This figure illustrates the transportation pathways of oil from the refinery to the market, highlighting the transition from non-mixing to mixing and finally to mixed oil during transport.
    }
    \label{fig:Mixed Oil}
\end{figure}

Oil products are one of the most critical energy sources for modern industry. Driven by technological progress, the primary mode of transporting oil products has gradually shifted toward pipeline transportation~\citep{Trench2001,Huang2021}, due to its low cost, minimal loss, high volume, and high safety~\citep{reniers2021}. In this multi-product pipeline transportation mode, various types of products, such as gasoline, home heating oil, and diesel fuel, are transported in the same pipeline without physical separation between batches~\citep{MirHassani2011}. Consequently, when different batches of products flow consecutively through the pipeline, an interface mixture forms between the adjacent batches. This contaminated segment is practically unusable, as it degrades the quality of pure products and causes substantial economic losses~\citep{Austin1963,Du2023a}. Figure~\ref{fig:Mixed Oil} illustrates the mixed oil formation phenomenon during transport. 
A common industrial approach to this issue involves predicting the length of the mixed oil in multi-product pipelines and isolating it accordingly \citep{du2025deeppipe}, so that improving the prediction accuracy directly reduces resource waste and enhances the economic efficiency of industrial production~\citep{Chen2021b,Patrachari2012}. Some researchers have employed hardware instruments, such as optical instruments and densitometers, to estimate mixed oil length, but these methods incur extremely high installation, operational, and maintenance costs, thus limiting their large-scale application in actual industrial settings.
Estimating the mixed oil length through physical mechanisms and machine learning methods therefore remains the most widely adopted approach, and at the initial stage of this research, scholars' efforts were largely devoted to exploring the industrial mechanism of mixed oil formation. \citet{Austin1963} proposed an equation to estimate mixed oil length, with empirical parameters estimated from large-scale industrial datasets. Despite their simplicity for quick predictions, these formulas are not universally reliable as their parameters vary by pipeline and environment~\citep{Li2023}. To improve the accuracy, \citet{baptista2001} refined the coefficients in Austin's prediction models, thereby enhancing their applicability. From the perspective of actual pipeline transportation, \citet{Wang2020} observed that near-wall adsorption effects are an important source of prediction deviations. \citet{Hamidi2023} introduced a two-dimensional numerical model that accounts for initial mixing, achieving more accurate results. However, the application of such methods in long-distance industrial pipelines remains challenging due to the heavy computational burden~\citep{Li2023}. 
With the rapid advancement of data science, researchers have begun to explore the use of machine and deep learning algorithms to construct effective models. These approaches aim to uncover the underlying relationships between pipeline features and mixed oil length, allowing mixed oil length to be predicted from pipeline features. \citet{Li2021} proposed a model based on partial least squares and deep neural networks to overcome the limitations of strong assumptions and the neglect of important features. In the same year, \citet{Chen2021b} integrated mechanistic models with the gradient boosting decision tree algorithm~\citep{Friedman2001} to develop a dual-driver model for predicting mixed oil length, which enhanced prediction accuracy. \citet{Yuan2021} developed hybrid modeling methods driven by both operation data and the mechanism based on the Austin-Palfrey equation. Following these studies, \citet{yuan2023} proposed a Student's t mixture regression method based on physical mechanisms, developing a more stable hybrid prediction model. 
\citet{yang2024confidence} proposed a method based on a diffusion model that learns the entire distribution rather than a single point estimate, thereby improving predictive performance.

\begin{figure}[pos=h]
    \centering
    \includegraphics[width=1\textwidth,trim=0 300 0 100,clip ]{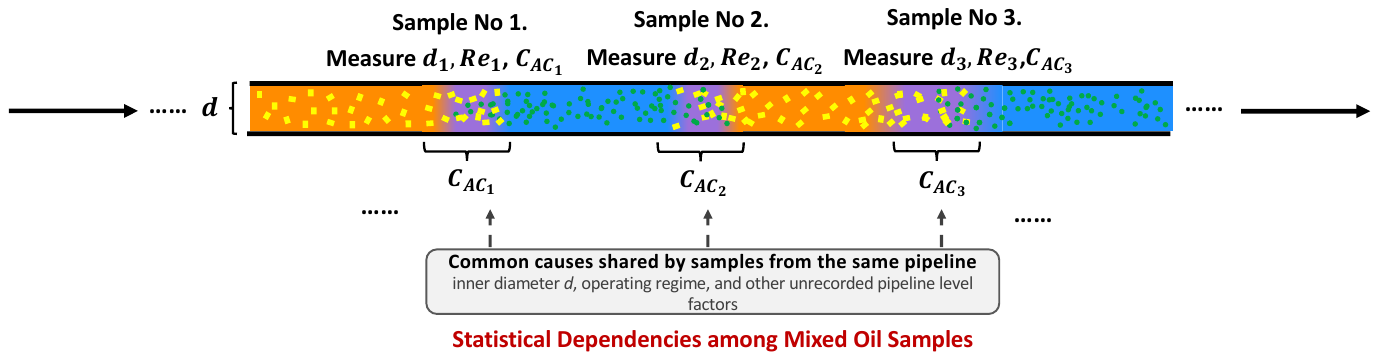}
        \caption{
        \textbf{The source of the statistical dependencies among mixed oil samples.} The figure depicts a pipeline from which the samples are collected. The samples share the conditions of that pipeline, and the arrows denote these shared common causes rather than physical interaction between the segments.
    }
    \label{fig:Inter-Sample Dependencies}
\end{figure}

However, these methods treat mixed oil samples as statistically independent observations. In practice, samples generated under shared pipeline and operating conditions may be statistically dependent, even when the oil segments do not physically interact. To illustrate how these shared conditions give rise to this dependence, we introduce an empirical formula. Specifically, we employ the Austin-Palfrey (AP) formula~\citep{Austin1963}, a mechanistic model for estimating the mixed oil length, to show where those shared conditions enter. We denote the inner diameter of the pipeline by $d$, the Reynolds number by $Re$, and the transportation distance by $L_0$. The formula is evaluated not at $L_0$ but at the total equivalent transportation distance $L_c$, which adds to $L_0$ the extra length of pipe that would by itself generate the mixed oil already present when transport begins, as Section~\ref{sec_distribution_shift} makes precise. Its coefficients are empirical and are tied to $d$ and $L_c$ being expressed in meters. According to the AP formula, the estimated mixed oil length is given by:
\begin{equation}
C_\mathrm{AP} =
\begin{cases}
11.75 \sqrt{dL_c}Re^{-0.1}, & Re \geq 10000e^{2.72\sqrt{d}}, \\
18384 \sqrt{dL_c}Re^{-0.9} e^{2.175\sqrt{d}}, & Re < 10000e^{2.72\sqrt{d}},
\end{cases}
\label{austin}
\end{equation}
where $C_{\text{AP}}$ is the mixed oil length estimated by the AP formula and $L_c$ is the total equivalent transportation distance, which absorbs the mixed oil already present when transport begins and is defined in Section~\ref{sec_distribution_shift}.
The AP formula shows that $C_\mathrm{AP}$ is determined by $O$, $d$, $L_0$, and $Re$, and since $C_\mathrm{AP}$ closely approximates the true mixed oil length $C_\mathrm{AC}$, the value of $C_\mathrm{AC}$ likewise depends on these four recorded features. The empirical coefficients of the AP formula are themselves known to vary with the pipeline and with the environment~\citep{Li2023}, so the recorded features do not determine $C_\mathrm{AC}$ completely, and the unrecorded conditions under which the samples are generated act as common causes of their targets. The samples therefore carry dependence from two sources, which Figure~\ref{fig:Inter-Sample Dependencies} illustrates using a single pipeline as an example.

\begin{itemize}
    \item[(i)] \textbf{Dependence carried by the recorded features}. Samples are collected from the same pipeline, sharing the same $d$. Their Reynolds numbers are also related, because $Re$ is calculated by 
    \begin{equation}
    Re = \frac{\rho v d}{\mu},
    \label{Re}
    \end{equation}
where \(\rho\) is the fluid density, \(v\) is the flow velocity, \(d\) is the inner diameter, and \(\mu\) is the fluid viscosity. Given that these samples share dependent values of \(\rho\), \(v\), and \(\mu\), and the same \(d\), the resulting Reynolds numbers \(Re_1\), \(Re_2\), and \(Re_3\) of three representative samples are also dependent. 
    \item[(ii)] \textbf{Dependence carried by the unrecorded factors}. The three samples pass through the same pipe wall and are handled under the same batch schedule, ambient temperature, and pumping practice. Wall roughness and near-wall adsorption~\citep{Wang2020}, the physicochemical contrast between adjacent products, and the operating regime therefore take common values across them, and none of these quantities appears among $O$, $d$, $L_0$, and $Re$. Acting as common causes of \(C_{\mathrm{AC}_1}\), \(C_{\mathrm{AC}_2}\), and \(C_{\mathrm{AC}_3}\), they leave the three targets dependent after the recorded features have been accounted for. This is the source that a model treating the samples as independent cannot use.
\end{itemize}
We next use the collected data to examine whether these statistical dependencies arise in practice. A pipeline is operated in sections that differ in inner diameter and transportation distance. Since samples from the same section share these conditions most closely, we use the section as the grouping unit. We quantify the dependence among these samples using the intraclass correlation coefficient (ICC), which measures the proportion of total variance attributable to differences between groups.
Across the 37 sections in our dataset, the ICC is $0.936$ for the measured mixed oil length $C_\mathrm{AC}$ and $0.494$ for the residual target $C_\Delta$ defined in Section~\ref{sec_distribution_shift}. After a linear fit on $O$, $d$, $L_0$ and $Re$ is subtracted from $C_\Delta$, the ICC of the residual remains $0.338$, as reported in Table~\ref{tab:icc}, so that a third of the residual variance is associated with differences between sections rather than with differences among samples of the same section. Since a linear adjustment cannot absorb a nonlinear conditional mean, this figure measures the clustering that a linear model leaves behind rather than conditional dependence in a strict sense, and we read it accordingly. An analysis in which each section is left out in turn points the same way. Predictions for a held out section carry a systematic offset whose square accounts on average for $32.9\%$ of the mean squared error of that section. Both analyses are diagnostics of association rather than identifications of a cause, and Appendix~\ref{loso} gives their definitions.

\begin{table}[pos=h]
\centering
\caption{Empirical within section dependence of the target, measured over the 37 pipeline sections of the dataset}
\label{tab:icc}
\begin{tabular}{lccc}
\toprule
Quantity & ICC & 95\% CI & $p$-value \\
\midrule
$C_{\mathrm{AC}}$                                        & 0.936 & [0.906, 0.961] & $<0.001$ \\
$C_{\Delta}$                                             & 0.494 & [0.389, 0.625] & $<0.001$ \\
Residual of $C_{\Delta}$ given $O$, $d$, $L_0$ and $Re$  & 0.338 & [0.249, 0.467] & $<0.001$ \\
\bottomrule
\end{tabular}
\end{table}

Graph representation learning offers a natural way to capture such statistical dependencies. A graph neural network represents entities as nodes and their relations as edges, and propagates information across connected nodes through message passing~\citep{kipf2016semi,hamilton2017inductive}. In particular, representing samples and features as the nodes of a graph lets a model learn from the relations encoded in its structure~\citep{you2020}. Connecting dependent samples through their shared features therefore turns the overlooked statistical dependencies into a usable learning signal.
Based on the above consideration, we propose a novel Sample Dependencies Incorporated Graph Representation Learning (SDI-GRL) model. This model explicitly incorporates the statistical dependencies among mixed oil samples by connecting them in a constructed graph, thereby enabling more effective use of information shared across dependent samples.
The main contributions of this paper are as follows:
\begin{itemize}

\item Characterization of statistical dependencies: We identify and quantify the statistical dependencies among mixed oil samples generated under shared pipeline and operating conditions using industrial data.

\item SDI-GRL model: We propose a mixed oil length prediction model based on graph representation learning. SDI-GRL represents samples and their shared features as nodes in a constructed graph and uses message passing to exchange information among statistically dependent samples. It therefore incorporates sample dependencies directly into the learning process and exploits information that methods assuming independent samples overlook.

\item Performance improvement: Compared with leading baselines, SDI-GRL reduces both RMSE and MAE by more than $10\%$.

\end{itemize}

The remainder of this paper is organized as follows. Section~\ref{sec:metho} describes the methodology, including SDI-GRL model structure, the training process and the novel design. Section~\ref{sec:case} presents experimental results, comparing our proposed model with eight baseline models, and provides a sensitivity analysis. Finally, Section~\ref{sec:conclu} concludes the paper and suggests future research directions. The main notations used in this paper are shown in Table~\ref{tab:notations}.

\begin{longtable}{
>{\centering\arraybackslash}p{0.28\textwidth}
p{0.64\textwidth}}
\caption{Notations used in this paper.}
\label{tab:notations}\\
\toprule
Notation & Meaning \\
\midrule
\endfirsthead

\multicolumn{2}{c}{\textbf{\tablename\ \thetable\ continued}}\\
\toprule
Notation & Meaning \\
\midrule
\endhead

\midrule
\multicolumn{2}{r}{Continued on the next page}\\
\endfoot

\bottomrule
\endlastfoot

\(n\), \(m\)
&
Sample size and number of pipeline features. The \(m\)-th feature is the prediction target \(C_{\Delta}\), which is never supplied to the graph as an input and is therefore carried by no edge of either graph.
\\

\(\alpha\), \(\beta\)
&
Indices running over samples and over features, respectively.
\\

\(\mathbf{X}\in\mathbb{R}^{n\times m}\)
&
Tabular data matrix, with \(\mathbf{X}_{\alpha\beta}\) the value of feature \(\beta\) for sample \(\alpha\).
\\

\(n_{\mathrm{train}}\), \(n_{\mathrm{test}}\)
&
Sizes of the training and test sets; \(n_{\mathrm{test}}=n-n_{\mathrm{train}}\).
\\

\(n_s\)
&
How many sample nodes an input graph carries: \(n_{\mathrm{train}}\) for \(\mathcal{G}_{\mathrm{train}}\), \(n_{\mathrm{test}}\) for \(\mathcal{G}_{\mathrm{test}}\).
\\

\(O\), \(d\), \(L_0\), \(Re\)
&
Initial mixed oil volume, pipeline inner diameter, transportation distance, and Reynolds number.
\\

\(C_0\), \(L_e\), \(L_c\)
&
Initial mixed oil length \(C_0=4O/(\pi d^{2})\); \(L_e\) is the distance over which the AP formula alone would generate \(C_0\), and \(L_c=L_0+L_e\) is the total equivalent distance at which Eq.~(\ref{austin}) is evaluated.
\\

\(\rho\), \(v\), \(\mu\)
&
Density, flow velocity, and viscosity of the transported fluid.
\\

\(C_{\mathrm{AC}}\)
&
Mixed oil length measured in the field.
\\

\(C_{\mathrm{AP}}\)
&
Estimate of the mixed oil length given by the Austin-Palfrey formula.
\\

\(C_{\Delta}\)
&
Residual left by that estimate, \(C_{\Delta}=C_{\mathrm{AC}}-C_{\mathrm{AP}}\), which serves as the prediction target of this paper.
\\

\(C_{\Delta,\alpha}\)
&
Residual of sample \(\alpha\); observed for \(\alpha\leq n_{\mathrm{train}}\) and to be predicted otherwise.
\\

\(\mathcal{G}_{\mathrm{train}}\), \(\mathcal{G}_{\mathrm{test}}\)
&
Bipartite graphs built separately from the training samples and from the test samples.
\\

\(\mathcal{V}_{\mathrm{train}}\), \(\mathcal{V}_{\mathrm{test}}\)
&
Node sets of the two graphs, each obtained by joining its own sample nodes with the shared feature nodes \(\mathcal{V}_{F}\).
\\

\(\mathcal{V}_{S,\mathrm{train}}\), \(\mathcal{V}_{S,\mathrm{test}}\)
&
Sample nodes alone, one for every training and every test sample.
\\

\(\mathcal{V}_{F}\)
&
Feature node set \(\{f_1,\ldots,f_m\}\), common to both graphs.
\\

\(\mathcal{E}_{\mathrm{train}}\), \(\mathcal{E}_{\mathrm{test}}\)
&
Their corresponding edge sets.
\\

\(s_{\alpha}\), \(f_{\beta}\)
&
Node standing for sample \(\alpha\), and node standing for feature \(\beta\).
\\

\(e_{\alpha\beta}\)
&
Value carried by the training edge between \(s_{\alpha}\) and \(f_{\beta}\), taken as \(\mathbf{X}_{\alpha\beta}\), for \(\beta=1,\ldots,m-1\).
\\

\(e'_{\alpha\beta}\)
&
Its test counterpart, which equals \(\mathbf{X}_{\alpha\beta}\) and is likewise defined for \(\beta=1,\ldots,m-1\) only.
\\

\(g_{\theta}\), \(g_{\theta^*}\)
&
SDI-GRL viewed as a mapping from an input graph to an edge matrix, before and after training.
\\

\(\theta\), \(\theta^*\)
&
All trainable parameters, and the values they take at the minimum of \(\mathcal{L}(\theta)\).
\\

\(L\), \(l\)
&
Depth of the message passing network, and the index of a single layer, \(l=1,\ldots,L\).
\\

\(d_V\), \(d_E\), \(d_M\)
&
Widths of the node, edge, and message embeddings.
\\

\(h_{s_\alpha}^{(l)}\), \(h_{f_\beta}^{(l)}\), \(h_{e_{\alpha\beta}}^{(l)}\)
&
Layer \(l\) embeddings on the training graph, for sample node \(s_\alpha\), feature node \(f_\beta\), and edge \(e_{\alpha\beta}\), respectively.
\\

\(h_{s_\alpha}^{\prime(l)}\), \(h_{f_\beta}^{\prime(l)}\), \(h_{e_{\alpha\beta}}^{\prime(l)}\)
&
Same quantities computed on the test graph, where the edge embedding now refers to \(e'_{\alpha\beta}\).
\\

\(\tilde h_{s_\alpha}\), \(\tilde h_{f_\beta}\)
&
Node representations passed to the readout, each obtained by concatenating the embeddings of the \(L\) layers and therefore of width \(d_VL\). Their test counterparts are \(\tilde h_{s_\alpha}'\) and \(\tilde h_{f_\beta}'\).
\\

\(\mathcal{N}(s_\alpha)\), \(\mathcal{N}(f_\beta)\)
&
Feature nodes adjacent to \(s_\alpha\) and sample nodes adjacent to \(f_\beta\) in the graph on which message passing is carried out.
\\

\(\mathbf{1}_m\)
&
All ones vector of length \(m\); every sample node starts from it.
\\

\(\mathbf{ONEHOT}_{\beta}\)
&
One hot vector in \(\mathbb{R}^{m}\) marking feature \(\beta\), used to initialize \(f_\beta\).
\\

\(m_{s_\alpha}^{(l)}\), \(m_{f_\beta}^{(l)}\)
&
Messages aggregated at layer \(l\) before the update of \(s_\alpha\) and of \(f_\beta\).
\\

\(W_1^{(l)},\ldots,W_5^{(l)}\)
&
Five learnable weight matrices belonging to layer \(l\).
\\

\(\mathrm{CONCAT}_V\)
&
Vertical concatenation of vectors.
\\

\(\mathrm{MAX}_s\), \(\mathrm{MAX}_f\)
&
Max aggregation operators: \(\mathrm{MAX}_s\) pools over neighboring sample nodes, \(\mathrm{MAX}_f\) over neighboring feature nodes.
\\

\(\mathcal{P}\), \(\Theta_{\mathcal{P}}\)
&
MLPNet applied to the concatenation of the layer embeddings in Eq.~(\ref{concat_layers}), together with the parameters it contains.
\\

\(\mathcal{H}\), \(\Theta_{\mathcal{H}}\)
&
Readout MLPNet at the output layer, together with the parameters it contains.
\\

\(\hat e_{\alpha\beta}\)
&
Reconstructed value of the training edge \(e_{\alpha\beta}\).
\\

\(\Omega_{\mathrm{drop}}\)
&
Edge dropout set. A proportion of the recorded feature edges of the training graph, resampled every epoch and deleted from the input, which regularizes the message passing.
\\

\(\mathcal{L}(\theta)\)
&
Training loss, the mean squared error between \(\hat e_{\alpha m}\) and \(C_{\Delta,\alpha}\) over the training samples.
\\

\(\hat C_{\Delta,\alpha}\)
&
Residual predicted for test sample \(\alpha\).
\\

\(\hat C_{\mathrm{AC},\alpha}\)
&
Mixed oil length finally reported, recovered by adding the AP estimate back: \(\hat C_{\mathrm{AC},\alpha}=\hat C_{\Delta,\alpha}+C_{\mathrm{AP},\alpha}\).
\\

\end{longtable}


\section{Methodology}
\label{sec:metho}
In this section, we first introduce the SDI-GRL framework, including the physical-domain knowledge integration and graph construction. Then, we describe the training and prediction procedures of SDI-GRL. Finally, we explain how SDI-GRL captures statistical dependencies among mixed oil samples through its graph architecture.

\subsection{Mitigating Distribution Shift with Physical Domain Knowledge}
\label{sec_distribution_shift}
Existing studies typically employ pipeline features to directly predict the mixed oil length $C_\mathrm{AC}$.
However, mixed oil datasets are prone to a distribution shift between the training and test data, meaning that the training samples and the test samples follow different distributions. Such a shift impairs the generalization ability of the model~\citep{yang2024confidence,moreno2012unifying}.
To mitigate this shift, we incorporate physical-domain knowledge into our model.  
Specifically, instead of predicting \( C_{\mathrm{AC}} \) directly, we predict the residual \( C_{\Delta} := C_{\mathrm{AC}} - C_{\mathrm{AP}} \). Since \( C_{\mathrm{AP}} \) already captures the dominant and physically explainable component of \( C_{\mathrm{AC}} \), the residual \( C_{\Delta} \) is smaller in magnitude and varies less across pipelines.
The learning target is therefore less sensitive to distribution shift, facilitating more robust model training and improved generalization.
We now specify how $C_{\mathrm{AP}}$ is evaluated. The AP formula of Eq.~(\ref{austin}) describes the mixed oil generated over a transportation distance, whereas part of the mixed oil is already present when transport begins. Following \citet{yang2024confidence}, we represent this initial amount by an extra length of pipe. An initial mixed oil volume $O$ occupies the initial mixed oil length $C_0=4O/(\pi d^{2})$, which the AP formula would itself produce over the equivalent distance
\begin{equation}
L_e=
\begin{cases}
\left(\dfrac{C_0 Re^{0.1}}{11.75\sqrt{d}}\right)^{2}, & Re \geq 10000e^{2.72\sqrt{d}}, \\[2ex]
\left(\dfrac{C_0 Re^{0.9}e^{-2.175\sqrt{d}}}{18384\sqrt{d}}\right)^{2}, & Re < 10000e^{2.72\sqrt{d}},
\end{cases}
\label{Le}
\end{equation}
so we evaluate Eq.~(\ref{austin}) at the total equivalent transportation distance $L_c=L_0+L_e$. Each branch of Eq.~(\ref{Le}) inverts the branch of Eq.~(\ref{austin}) that applies at the same Reynolds number, and a sample is assigned to a branch by its own value of $Re$, so that $C_0$ and $C_{\mathrm{AP}}$ are always generated by one and the same mechanism.

\subsection{Problem Formulation}
We let $\mathbf{X} \in \mathbb{R}^{n \times m}$ be a tabular data matrix consisting of $n$ samples and $m$ pipeline features, where the $\beta$-th feature of the $\alpha$-th sample is denoted as $\mathbf{X}_{\alpha \beta}$. Specifically, the $m$-th feature is defined as the prediction target $C_{\Delta}$ introduced in Section~\ref{sec_distribution_shift}. We divide the samples into a training set and a test set: the first
$n_{\mathrm{train}}$ samples constitute the training set and the remaining
$n_{\mathrm{test}}=n-n_{\mathrm{train}}$ samples constitute the test set.
The target $C_{\Delta,\alpha}$ is observed for $\alpha\leq n_{\mathrm{train}}$
and is to be predicted for $\alpha>n_{\mathrm{train}}$. We consider the mixed oil length prediction problem, where the goal is to learn a prediction model from the training set 
\(\{(\mathbf{X}_{\alpha,1:m-1}, C_{\Delta,\alpha})\}_{\alpha=1}^{n_{\mathrm{train}}}\),
and apply the trained model to predict the unobserved target values 
\(C_{\Delta,\alpha}\) for the test set 
\(\{\mathbf{X}_{\alpha,1:m-1}\}_{\alpha=n_{\mathrm{train}}+1}^{n}\).
\subsection{Sample Dependencies Incorporated Graph Representation Learning Model}
\label{sec:sdigrl}
We propose the Sample Dependencies Incorporated Graph Representation Learning (SDI-GRL) model to predict the mixed oil length for the test set. This model is capable of indirectly and effectively capturing statistical dependencies among mixed oil samples, which existing methods overlook. We present the detailed methodology of SDI-GRL consisting of three main steps.

\begin{itemize}
\item \textbf{Step 1: Represent the training data as a training graph.}
We represent the training data
\(\{\mathbf{X}_{\alpha,1:m}\}_{\alpha=1}^{n_{\mathrm{train}}}\)
as a training graph
\(\mathcal{G}_{\mathrm{train}}=(\mathcal{V}_{\mathrm{train}},\mathcal{E}_{\mathrm{train}})\).
Specifically, the node set
\(\mathcal{V}_{\mathrm{train}}\) consists of the sample node set
\(\mathcal{V}_{S,\mathrm{train}}\) and the feature node set
\(\mathcal{V}_{F}\), i.e.,
\(\mathcal{V}_{\mathrm{train}}=\mathcal{V}_{S,\mathrm{train}}\cup\mathcal{V}_{F}\),
where
\(\mathcal{V}_{S,\mathrm{train}}=\{s_1,s_2,\ldots,s_{n_{\mathrm{train}}}\}\)
represents the training mixed oil samples and
\(\mathcal{V}_{F}=\{f_1,f_2,\ldots,f_m\}\)
represents the \(m\) pipeline features.
The edge set is defined as
$
\mathcal{E}_{\mathrm{train}}
=
\{(s_\alpha,f_\beta,e_{\alpha\beta})
\mid
1\leq\alpha\leq n_{\mathrm{train}},
1\leq\beta\leq m-1\},
$
where \(e_{\alpha\beta}\) denotes the edge value between the sample node \(s_\alpha\) and the feature node \(f_\beta\), and is taken as \(\mathbf{X}_{\alpha\beta}\). The index \(\beta\) stops at \(m-1\), such that the target column never enters the graph as an input. The target node \(f_m\) therefore carries no incident edge, and the observed values \(C_{\Delta,\alpha}\) serve exclusively as the regression targets of Eq.~(\ref{loss}). Since the target column is likewise absent from the test graph of Step 3, the model receives the same kind of input during training as it does at inference.
\item \textbf{Step 2: Train SDI-GRL model.}
The SDI-GRL model defines a parameterized mapping \(g_\theta:\mathcal{G}\rightarrow\mathbb{R}^{n_s\times m}\), where \(\mathcal{G}\) denotes the input graph, \(\theta\) denotes the trainable parameters of the model, and \(n_s\) denotes the number of sample nodes in the input graph, i.e., \(n_s=n_{\mathrm{train}}\) for \(\mathcal{G}_{\mathrm{train}}\). Taking \(\mathcal{G}_{\mathrm{train}}\) as input, SDI-GRL learns graph representations through graph neural network message passing, resulting in the optimized parameters \(\theta^*\) and the trained model \(g_{\theta^*}\). The detailed training procedure is provided in Section~\ref{sec:Train}.
\item \textbf{Step 3: Apply the trained model to predict the mixed oil length.}
After training, we construct a test graph
\(\mathcal{G}_{\mathrm{test}}=(\mathcal{V}_{\mathrm{test}},\mathcal{E}_{\mathrm{test}})\)
using the test samples as follows. Specifically, the node set
\(\mathcal{V}_{\mathrm{test}}\) consists of the sample node set
\(\mathcal{V}_{S,\mathrm{test}}\) and the feature node set
\(\mathcal{V}_{F}\), i.e.,
\(\mathcal{V}_{\mathrm{test}}=\mathcal{V}_{S,\mathrm{test}}\cup\mathcal{V}_{F}\), where the sample node set is defined as
\(\mathcal{V}_{S,\mathrm{test}}=\{s_{n_{\mathrm{train}}+1},\ldots,s_n\}\).
The edge set is constructed as
$
\mathcal{E}_{\mathrm{test}}
=
\{(s_\alpha,f_\beta,e'_{\alpha\beta})
\mid
n_{\mathrm{train}}+1\leq\alpha\leq n,
1\leq\beta\leq m-1\},
$
where \(e'_{\alpha\beta}=\mathbf{X}_{\alpha\beta}\) for \(1\leq\beta\leq m-1\). The target node \(f_m\) again carries no incident edge, which represents the unobserved target of a test sample by the absence of an edge rather than by a substitute value. This matters because the scaling of Section~\ref{sec_dataset} places the recorded values on a common numerical range, such that any value assigned in place of a missing one would be indistinguishable from a legitimate observation. Figure~\ref{fig:traintest} of Appendix~\ref{flowchart} sets the two graphs side by side.
The trained model \(g_{\theta^*}\) is then applied to \(\mathcal{G}_{\mathrm{test}}\) to predict the mixed oil length. The detailed implementation of the prediction procedure is provided in Section~\ref{predict}.
\end{itemize}
\begin{algorithm}[htb]
\caption{Mixed oil length prediction based on SDI-GRL}
\label{alg:1}
\begin{algorithmic}[1]
\Statex \textbf{Input:} Dataset \(\mathbf{X}\), number of GNN layers \(L\), number of epochs \(T\)
\Statex \textbf{Output:} Predicted mixed oil length \(\hat{C}_{\mathrm{AC}}\)
\State \textbf{Step 1: Represent the training data as a training graph}
\State Construct training graph \(\mathcal{G}_{\mathrm{train}}\) according to Section~\ref{sec:sdigrl}
\State \textbf{Step 2: Train SDI-GRL model}
\State  Initialize node embeddings \(h_{s_\alpha}^{(0)}, h_{f_\beta}^{(0)}\) for \(\mathcal{G}_{\mathrm{train}}\) through Eqs.~(\ref{init_sample})-(\ref{init_feature})
\For{\(t=1\) to \(T\)}
\State  Sample the edge dropout set \(\Omega_{\mathrm{drop}}\), delete the edges it indexes, and initialize the embeddings \(h_{e_{\alpha\beta}}^{(0)}\) of the retained edges through Eq.~(\ref{init_edge})
\For{\(l=1\) to \(L\)}
    \State Update node and edge embeddings \(h_{s_\alpha}^{(l)}, h_{f_\beta}^{(l)}, h_{e_{\alpha\beta}}^{(l)}\) through Eqs.~(\ref{msx})-(\ref{esxl})
\EndFor
\State Compute edge predictions \(\hat e_{\alpha\beta}\) by Eq.~(\ref{output})
\State Minimize \(\mathcal{L}(\theta)\) by Eq.~(\ref{loss})
\EndFor
\State Obtain \(\theta^*\) and trained model \(g_{\theta^*}\)
\State \textbf{Step 3: Apply the trained model to predict the mixed oil length}
\State Construct test graph \(\mathcal{G}_{\mathrm{test}}\) according to Section~\ref{sec:sdigrl}
\State Initialize node and edge embeddings \(h_{s_\alpha}'^{(0)}, h_{f_\beta}'^{(0)}, h_{e_{\alpha\beta}}'^{(0)}\) for \(\mathcal{G}_{\mathrm{test}}\) through Eqs.~(\ref{init_sample_test})-(\ref{init_edge_test})
\State Perform forward computation on \(\mathcal{G}_{\mathrm{test}}\) with fixed parameters \(\theta^*\) through Eqs.~(\ref{msx})-(\ref{esxl})
\State Obtain \(\hat{C}_{\Delta,\alpha}\) through Eq.~(\ref{predict_delta})
\State Predict the mixed oil length \(\hat{C}_{\mathrm{AC}}\) through Eq.~(\ref{predict_ac})
\end{algorithmic}
\end{algorithm}
\subsection{The Detailed Training of SDI-GRL}
\label{sec:Train}
Next, we present the detailed training procedure of the SDI-GRL model described in Step 2. The training of SDI-GRL is implemented using an \(L\)-layer graph neural network. Since conventional neural networks cannot directly operate on graph-structured data, we first initialize the node and edge representations of \(\mathcal{G}_{\mathrm{train}}\) as embeddings, following existing graph representation learning approaches~\citep{kipf2016semi,grover2016node2vec}. The initialized embeddings are then updated through \(L\) layers of graph neural network message passing, enabling SDI-GRL to effectively capture the structural dependencies among mixed oil samples and pipeline features~\citep{abu2018watch}. The detailed procedures are described as follows.
\begin{itemize}
\item \textbf{(I): Define and Initialize Embeddings.}
We first represent the nodes and edges of \(\mathcal{G}_{\mathrm{train}}\) as embeddings~\citep{hamilton2017inductive}. Specifically, we let the sample node embedding be \(h^{(l)}_{s_\alpha}\), \(\alpha=1,\ldots,n_{\mathrm{train}}\), the feature node embedding be \(h^{(l)}_{f_\beta}\), \(\beta=1,\ldots,m\), and the edge embedding be \(h^{(l)}_{e_{\alpha\beta}}\), \(\alpha=1,\ldots,n_{\mathrm{train}},\beta=1,\ldots,m-1\), where \(l\) denotes the layer of the graph neural network. We initialize them as follows:
\begin{equation}
\label{init_sample}
h^{(0)}_{s_\alpha}=\mathbf{1}_m,
\quad
\alpha=1,\ldots,n_{\mathrm{train}},
\end{equation}
\begin{equation}
\label{init_feature}
h^{(0)}_{f_\beta}=\mathbf{ONEHOT}_{\beta},
\quad
\beta=1,\ldots,m,
\end{equation}
\begin{equation}
\label{init_edge}
h^{(0)}_{e_{\alpha\beta}}=
e_{\alpha\beta},
\quad
(\alpha,\beta)\notin\Omega_{\mathrm{drop}},
\end{equation}
where \(\mathbf{1}_m\) is an \(m\)-dimensional vector of ones, and \(\mathbf{ONEHOT}_{\beta}\) is an \(m\)-dimensional vector with only the \(\beta\)-th element being 1 and all others 0.  Here \(\Omega_{\mathrm{drop}}\) is the edge dropout set, formed at the beginning of every epoch by drawing a proportion \(r_{\mathrm{drop}}\) of the edges of \(\mathcal{E}_{\mathrm{train}}\) at random. The edges it indexes are deleted for the duration of that epoch, which leaves the epoch graph \(\mathcal{G}^{\mathrm{ret}}_{\mathrm{train}}=(\mathcal{V}_{\mathrm{train}},\mathcal{E}^{\mathrm{ret}}_{\mathrm{train}})\) with edge set
$
\mathcal{E}^{\mathrm{ret}}_{\mathrm{train}}
=
\{(s_\alpha,f_\beta,e_{\alpha\beta})\in\mathcal{E}_{\mathrm{train}}
\mid
(\alpha,\beta)\notin\Omega_{\mathrm{drop}}\}.
$
Since \(\mathcal{E}_{\mathrm{train}}\) contains no target edge, the deletion applies to the recorded features alone, and its purpose is regularization rather than supervision. Deletion removes the adjacency itself rather than the value an edge carries, so that the neighborhood of a node differs from one epoch to the next. We write \(\mathcal{N}(s_\alpha)\) and \(\mathcal{N}(f_\beta)\) for the feature nodes adjacent to \(s_\alpha\) and the sample nodes adjacent to \(f_\beta\) in the graph on which message passing is carried out. On the epoch graph these are \(\mathcal{N}(s_\alpha)=\{\beta\leq m-1\mid(\alpha,\beta)\notin\Omega_{\mathrm{drop}}\}\) and \(\mathcal{N}(f_\beta)=\{\alpha\mid(\alpha,\beta)\notin\Omega_{\mathrm{drop}}\}\), whereas on the test graph of Section~\ref{predict}, where no edge is deleted, \(\mathcal{N}(s_\alpha)\) runs over all \(m-1\) recorded features and \(\mathcal{N}(f_\beta)\) over all test samples. We adopt throughout the convention that an aggregation over an empty neighborhood returns the zero vector of the appropriate width, such that Eqs.~(\ref{msx}) and~(\ref{msb}) remain defined whenever a node retains no edge. Two cases arise. Since \(\mathcal{N}(f_m)=\emptyset\) in both graphs, we have \(m^{(l)}_{f_m}=\mathbf{0}\) at every layer, whence Eq.~(\ref{hsb}) updates \(f_m\) from its own previous embedding alone. The resulting \(h^{(l)}_{f_m}\) is of width \(d_V\), as for every other feature node, and is a function of \(\theta\) and of no sample, such that \(f_m\) supplies the readout with a learned query rather than with a target value. A sample node retains no edge only if the dropout removes all \(m-1\) of its edges, an event of probability \(r_{\mathrm{drop}}^{m-1}\), which equals \(0.0016\) at the ratio reported in Appendix~\ref{hyper}, and Eq.~(\ref{hsl}) then updates that node from its own previous embedding in the same way. \
\item \textbf{(II): Update the Embeddings through Message Passing.}
For layer \(l\) (\(l\geq1\)), we update the sample node embeddings \(h^{(l)}_{s_\alpha}\), feature node embeddings \(h^{(l)}_{f_\beta}\), and edge embeddings \(h^{(l)}_{e_{\alpha\beta}}\), respectively.
\begin{itemize}
\item Update sample node embeddings:
In order to update \(h^{(l)}_{s_\alpha}\), we first define the sample message \(m_{s_\alpha}^{(l)}\) as follows:
\begin{equation}
m_{s_\alpha}^{(l)}
=
\operatorname{MAX}_f
\left\{
\text{ReLU}
\left(
W_1^{(l)}
\cdot
\mathrm{CONCAT}_V
[
h_{s_\alpha}^{(l-1)},
h_{e_{\alpha\beta}}^{(l-1)}
]
\right)
\mid
\beta\in\mathcal{N}(s_\alpha)
\right\},
\alpha=1,\ldots,n_{\mathrm{train}},
\label{msx}
\end{equation}
where \(W_1^{(l)}\) is a learnable parameter matrix, \(\mathrm{CONCAT}_V\) denotes vertical vector concatenation, and \(\operatorname{MAX}_f\) denotes max aggregation over neighboring feature nodes. We then update \(h^{(l)}_{s_\alpha}\) as follows:
\begin{equation}
h_{s_\alpha}^{(l)}
=
\text{ReLU}
\left(
W_2^{(l)}
\cdot
\mathrm{CONCAT}_V
[
h_{s_\alpha}^{(l-1)},
m_{s_\alpha}^{(l)}
]
\right),
\alpha=1,\ldots,n_{\mathrm{train}},
\label{hsl}
\end{equation}
where \(W_2^{(l)}\) is a learnable parameter matrix.
\item Update feature node embeddings:
Similarly, we introduce the feature message \(m_{f_\beta}^{(l)}\) as follows:
\begin{equation}
m_{f_\beta}^{(l)}
=
\operatorname{MAX}_s
\left\{
\text{ReLU}
\left(
W_3^{(l)}
\cdot
\mathrm{CONCAT}_V
[
h_{f_\beta}^{(l-1)},
h_{e_{\alpha\beta}}^{(l-1)}
]
\right)
\mid
\alpha\in\mathcal{N}(f_\beta)
\right\},
\beta=1,\ldots,m,
\label{msb}
\end{equation}
where \(W_3^{(l)}\) is a learnable parameter matrix, and \(\operatorname{MAX}_s\) denotes max aggregation over neighboring sample nodes. We then update \(h^{(l)}_{f_\beta}\) as follows:
\begin{equation}
h_{f_\beta}^{(l)}
=
\text{ReLU}
\left(
W_4^{(l)}
\cdot
\mathrm{CONCAT}_V
[
h_{f_\beta}^{(l-1)},
m_{f_\beta}^{(l)}
]
\right),
\beta=1,\ldots,m,
\label{hsb}
\end{equation}
where \(W_4^{(l)}\) is a learnable parameter matrix.
\item Update edge embeddings:
Combining Eq.~(\ref{hsl}) and Eq.~(\ref{hsb}), we update \(h^{(l)}_{e_{\alpha\beta}}\) as follows:
\begin{equation}
h^{(l)}_{e_{\alpha \beta}}
=
\text{ReLU}
\left(
W_5^{(l)}
\cdot
\mathrm{CONCAT}_V
[
h^{(l-1)}_{e_{\alpha\beta}},
h_{s_\alpha}^{(l)},
h_{f_\beta}^{(l)}
]
\right),
\alpha=1,\ldots,n_{\mathrm{train}},
\beta\in\mathcal{N}(s_\alpha),
\label{esxl}
\end{equation}
where \(W_5^{(l)}\) is a learnable parameter matrix. It can be observed from Eq.~(\ref{esxl}) that \(h^{(l)}_{e_{\alpha\beta}}\) effectively integrates the information from \(h^{(l)}_{s_\alpha}\) and \(h^{(l)}_{f_\beta}\), thereby bridging mixed oil samples and pipeline features. Such integration facilitates the incorporation of dependencies among mixed oil samples, which will be discussed in Section~\ref{sec:novel}.
\end{itemize}
The above message passing operations are iteratively applied for \(l=1,\ldots,L\) following Eqs.~(\ref{msx})--(\ref{esxl}) until the \(L\)-th layer.
\item \textbf{(III): Edge Prediction at the Output Layer.}
After \(L\) layers of message passing, we gather the embeddings produced at every layer into the node representations
\begin{equation}
\label{concat_layers}
\tilde h_{s_\alpha}
=
\mathcal{P}
\left(
\mathrm{CONCAT}_V
[
h_{s_\alpha}^{(1)},\ldots,h_{s_\alpha}^{(L)}
]
\right),
\qquad
\tilde h_{f_\beta}
=
\mathcal{P}
\left(
\mathrm{CONCAT}_V
[
h_{f_\beta}^{(1)},\ldots,h_{f_\beta}^{(L)}
]
\right),
\end{equation}
where \(\mathcal{P}\) is a learnable MLPNet shared by the sample and the feature nodes that preserves the width \(d_VL\). Concatenating over the layers lets the representation formed at a shallow layer reach the output without having to pass through the deeper ones. The output of the model
\(g_\theta(\mathcal{G}_{\mathrm{train}})\) is then the edge matrix
$
g_\theta(\mathcal{G}_{\mathrm{train}})
=
\left[
\hat e_{\alpha\beta}
\right]_{\alpha=1,\ldots,n_{\mathrm{train}},
\beta=1,\ldots,m}
\in
\mathbb{R}^{n_{\mathrm{train}}\times m}.
$
Specifically, each edge value corresponding to the \(\alpha\)-th mixed oil sample and the \(\beta\)-th pipeline feature is obtained by
\begin{equation}
\label{output}
\hat{e}_{\alpha\beta}
=
\mathcal{H}
\left(
\mathrm{CONCAT}_V
[
\tilde h_{s_\alpha},
\tilde h_{f_\beta}
]
\right),
\quad
\alpha=1,\ldots,n_{\mathrm{train}},
\ \beta=1,\ldots,m ,
\end{equation}
where \(\mathcal{H}\) is a learnable multilayer perceptron network (MLPNet)~\citep{you2020}.
\item \textbf{(IV): Construct the Training Objective and Optimize Parameters.}
Steps (II) and (III) constitute one training epoch of SDI-GRL. A fixed number of epochs is performed, as reported in Appendix~\ref{hyper}. In each epoch, the edges in \(\Omega_{\mathrm{drop}}\) are deleted from the training graph through Eq.~(\ref{init_edge}) and thus take no part in message passing, which prevents the model from relying on any one recorded value~\citep{srivastava2014dropout}. The edge dropout ratio \(r_{\mathrm{drop}}\) is likewise reported in Appendix~\ref{hyper}. Since the target column is absent from \(\mathcal{E}_{\mathrm{train}}\), the loss is formed on the target edges alone, and it reads
\begin{equation}
\label{loss}
\mathcal{L}(\theta)
=
\frac{1}{n_{\mathrm{train}}}
\sum_{\alpha=1}^{n_{\mathrm{train}}}
(\hat{e}_{\alpha m}-C_{\Delta,\alpha})^2,
\end{equation}
where
$
\theta=
\{W_1^{(l)},\ldots,W_5^{(l)}\}_{l=1}^{L}
\cup
\Theta_{\mathcal{P}}
\cup
\Theta_{\mathcal{H}}
$ denotes the model parameters, in which \(W_1^{(l)}\) to \(W_5^{(l)}\) are the weight matrices of layer \(l\), \(\Theta_{\mathcal{P}}\) denotes the parameters of the MLPNet \(\mathcal{P}\) in Eq.~(\ref{concat_layers}), and \(\Theta_{\mathcal{H}}\) denotes the parameters of the MLPNet \(\mathcal{H}\) in Eq.~(\ref{output}). We aim to optimize the parameters to achieve convergence by minimizing the loss function through backpropagation and gradient-based optimization~\citep{rumelhart1986learning,kingma2014adam}. Specifically, the optimized parameters are obtained as
\begin{equation}
\label{theta}
\theta^*
=
\arg\min_{\theta}\mathcal{L}(\theta),
\end{equation}
where \(\theta^*=\{W_1^{*(l)},W_2^{*(l)},W_3^{*(l)},W_4^{*(l)},W_5^{*(l)}\}_{l=1}^{L}\cup\Theta_{\mathcal{P}}^*\cup\Theta_{\mathcal{H}}^*\) denotes the optimized model parameters. The resulting parameters \(\theta^*\) yield the trained SDI-GRL model \(g_{\theta^*}\).
\end{itemize}
\subsection{Predict the Mixed Oil Length}
\label{predict}
After obtaining the trained SDI-GRL model \(g_{\theta^*}\), we apply the optimized parameters \(\theta^*\) to the constructed test graph \(\mathcal{G}_{\mathrm{test}}\) and perform forward computation to predict the mixed oil length. Specifically, the test graph is initialized with node and edge embeddings \(h_{s_\alpha}^{\prime(0)}, h_{f_\beta}^{\prime(0)}, h_{e_{\alpha\beta}}^{\prime(0)}\), where \(\alpha=n_{\mathrm{train}}+1,\ldots,n\) denotes the test sample nodes and \(\beta=1,\ldots,m-1\) indexes the edges that carry the recorded features. The initialization is defined as
\begin{equation}
\label{init_sample_test}
h_{s_\alpha}^{\prime(0)}
=
\mathbf{1}_m,
\quad
\alpha=n_{\mathrm{train}}+1,\ldots,n,
\end{equation}
\begin{equation}
\label{init_feature_test}
h_{f_\beta}^{\prime(0)}
=
\mathbf{ONEHOT}_{\beta},
\quad
\beta=1,\ldots,m,
\end{equation}
\begin{equation}
\label{init_edge_test}
h_{e_{\alpha\beta}}^{\prime(0)}
=
e_{\alpha\beta}^{\prime},
\quad
\alpha=n_{\mathrm{train}}+1,\ldots,n,
\ \beta=1,\ldots,m-1 .
\end{equation}
The initialized embeddings are then updated through \(L\) layers of message passing following Eqs.~(\ref{msx})-(\ref{esxl}) with the fixed optimized parameters \(\theta^*\). During this forward computation, all sample-node indices in the message-passing operations are replaced by \(\alpha=n_{\mathrm{train}}+1,\ldots,n\) to match the test graph, and no edge is deleted, so \(\mathcal{N}(s_\alpha)\) runs over all \(m-1\) recorded features and \(\mathcal{N}(f_\beta)\) over all test samples. As on the training graph, \(f_m\) has no incident edge and is therefore updated from its own embedding alone under the convention of Section~\ref{sec:Train}. Collecting the resulting embeddings layer by layer as in Eq.~(\ref{concat_layers}) gives the test node representations \(\tilde h_{s_\alpha}'\) and \(\tilde h_{f_\beta}'\). According to Eq.~(\ref{output}), the predicted value of the \(m\)-th pipeline feature for each test sample is obtained as
\begin{equation}
\label{predict_delta}
\hat{C}_{\Delta,\alpha}
=
\mathcal{H}_{\Theta_{\mathcal{H}}^{*}}
\left(
\mathrm{CONCAT}_V
[
\tilde h_{s_\alpha}',
\tilde h_{f_m}'
]
\right),
\quad
\alpha=n_{\mathrm{train}}+1,\ldots,n,
\end{equation}
where $\mathcal{H}_{\Theta_{\mathcal{H}}^{*}}$ denotes the readout MLPNet $\mathcal{H}$
equipped with its optimized parameters $\Theta_{\mathcal{H}}^{*}\subset\theta^{*}$
obtained from Eq.~(\ref{theta}). Finally, based on the physical relationship introduced in Section~\ref{sec_distribution_shift}, the predicted mixed oil length is obtained as
\begin{equation}
\label{predict_ac}
\hat{C}_{\mathrm{AC},\alpha}
=
\hat{C}_{\Delta,\alpha}+C_{\mathrm{AP},\alpha},
\quad
\alpha=n_{\mathrm{train}}+1,\ldots,n .
\end{equation}
We evaluate Eqs.~(\ref{init_sample_test}) to~(\ref{predict_ac}) on \(\mathcal{G}_{\mathrm{test}}\) as a whole rather than sample by sample, such that the prediction \(\hat C_{\Delta,\alpha}\) returned for a test sample depends on the recorded features of every other test sample through the feature nodes they share. We therefore describe inference in SDI-GRL as transductive, in the sense that a prediction is conditioned on the composition of the test cohort and not on the sample in isolation. Such conditioning is precisely what a predictor acting on one row at a time cannot perform, and Section~\ref{sec:novel} describes the pathway along which SDI-GRL makes use of it. We emphasize that it does not constitute leakage, since only the recorded features of the test samples enter the computation, the parameters \(\theta^*\) are fixed before any test sample is seen, and no target value is available to the model at any point of the forward pass. Such a protocol is moreover the one that deployment presents, since the batch records of a pipeline reach an operator together rather than one at a time.

\subsection{Mechanism of Statistical Dependency Incorporation in SDI-GRL}
\label{sec:novel}
\begin{figure}[pos=h]
    \centering
   \includegraphics[width=1\textwidth,trim=0 250 0 0,clip ]{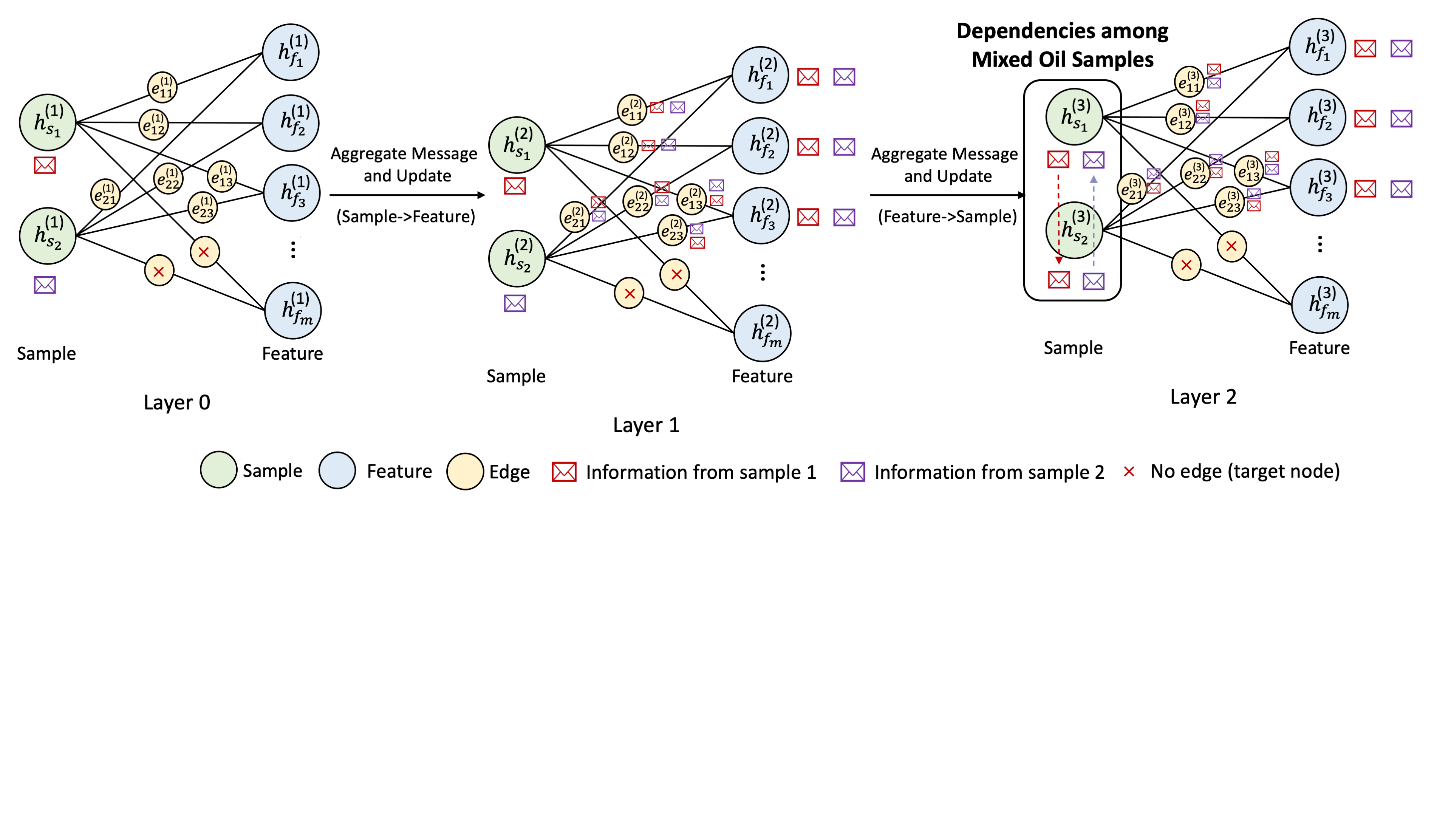}
        \caption{
        \textbf{Novel design of SDI-GRL.} The figure follows two mixed oil samples and the recorded feature nodes through the first two message passing layers, the leftmost panel showing the initialization and each subsequent panel the state reached after the update named on the arrow that precedes it. At layer 0 the information of a sample resides with that sample alone. The first layer carries it to the feature nodes, where the values of both samples are pooled. The second layer returns the pooled summary along the edges, so that \(h^{(2)}_{s_1}\) carries a contribution originating from \(s_2\) although the two sample nodes are never adjacent. The target node \(f_m\) has no incident edge, which the crosses mark.
    }
    \label{fig:novel design}
\end{figure}
As established in Section~\ref{sec:intr}, mixed oil samples generated under shared conditions remain statistically dependent once the recorded features have been accounted for. A model that maps a single row of $\mathbf{X}$ to a single target is unable to respond to such dependence, since its output is by construction a function of that row alone, and the remaining rows of the cohort leave it unchanged. The construction of Section~\ref{sec:sdigrl} removes this restriction, in that the $m-1$ feature nodes that carry the recorded values are common to every sample. Two sample nodes $s_\alpha$ and $s_{\alpha'}$ are never adjacent, yet they are joined by $m-1$ paths of length two, one through each of those feature nodes, so that a representation formed for one sample is conditioned on the values taken by the others. We emphasize that no target enters this pathway, since Section~\ref{sec:sdigrl} withholds the target column from both graphs. What the pathway conveys is the composition of the cohort in the recorded features, which serves as a proxy for the operating context that the individual record does not disclose.

Message passing turns these paths into an actual transfer of information, and Figure~\ref{fig:novel design} traces the transfer over the first layers for two samples and the $m-1$ recorded features. The layers of one round act in opposite directions. At the first layer, Eq.~(\ref{msb}) aggregates over $\mathcal{N}(f_\beta)$, so that $h^{(1)}_{f_\beta}$ summarizes the values taken on feature $\beta$ by every sample incident on $f_\beta$, and the information of $s_1$ and that of $s_2$ are pooled at the feature node accordingly. The sample update of the same layer draws only on $h^{(0)}_{s_\alpha}$ and the initial edge embeddings, so that $h^{(1)}_{s_\alpha}$ still rests on the values of $s_\alpha$ alone. At the second layer the direction is reversed. Eq.~(\ref{esxl}) has by then written $h^{(1)}_{f_\beta}$ into the edge embeddings, and Eq.~(\ref{msx}) aggregates these over $\mathcal{N}(s_\alpha)$, whence $h^{(2)}_{s_1}$ carries a contribution originating from $s_2$ although the two sample nodes are never adjacent. A single round of exchange between two samples therefore consumes two message passing updates, from which the role of the depth $L$ follows. In contrast to a feed-forward network, in which depth governs the number of successive transformations applied to a fixed input, $L$ here governs the number of rounds of exchange among dependent samples, and it is on this account that $L$ need not be tied to the number of features. Iterating the two stages to layer $L$ and collecting the layer embeddings by Eq.~(\ref{concat_layers}) yields $\tilde h_{s_\alpha}$, which the readout of Eq.~(\ref{output}) maps to $\hat e_{\alpha m}$. The prediction obtained for a sample is accordingly a function of that sample together with the cohort to which it belongs, and it is in this sense, namely as a transductive aggregation over the cohort rather than as an explicit encoding of section membership, that the statistical dependence among the mixed oil samples is incorporated into SDI-GRL.
\subsection{Mitigating Over-smoothing at Increased Depth}
A concern with repeated aggregation is that the embeddings of distinct nodes may be driven toward a common value, an effect referred to as over-smoothing. Four features of our construction work against this tendency, although none of them excludes it as a matter of principle. First, each update in Eqs.~(\ref{msx}) to~(\ref{esxl}) concatenates the previous embedding of a node with the incoming message before the linear map, so that the older representation remains available to the next layer, even though the learned weights may still discard it and the concatenation is therefore weaker than an identity path. Second, the aggregation is a maximum rather than a mean, which retains the salient component of a neighborhood instead of its center, although a maximum over similar neighborhoods will itself converge. Third, the edges indexed by $\Omega_{\mathrm{drop}}$ are deleted rather than assigned a null value, so that the adjacency, and with it the neighborhood over which a node aggregates, is resampled at every epoch. Fourth, and most decisively, the readout of Eq.~(\ref{output}) receives $\tilde h_{s_\alpha}$ and $\tilde h_{f_\beta}$ rather than the embeddings of the last layer alone, whence the representations formed at the shallow layers, on which the aggregation has acted the fewest times, reach the prediction directly however much the deeper layers smooth. Taken together, these features suggest that the depth $L$ need not be kept small on account of over-smoothing alone. Since the argument is architectural rather than a proof, we examine the effect of $L$ empirically in Section~\ref{ablation}.

\section{Case Study}
\label{sec:case}

\subsection{Dataset}
\label{sec_dataset}
The mixed oil dataset used in this study was supplied by an industrial pipeline operator under a confidentiality agreement and was extracted from the historian of the operator's \textit{Supervisory Control And Data Acquisition (SCADA)} system, the class of industrial control system described by \citet{boyer2004scada}. The dataset includes \(n=2309\) samples with \(m=5\) features: the initial mixed oil volume \(O\), the pipeline's inner diameter \(d\), the transportation distance \(L_0\), the Reynolds number \(Re\) and the prediction target $C_\Delta$. We set the training sample size to \(n_{\mathrm{train}} = 1781\) and the test sample size to \(n_{\mathrm{test}} = 528\), where the 1781 samples come from the first two pipelines and the 528 samples from another pipeline. These data are then fed into Algorithm~\ref{alg:1} to estimate the mixed oil length, with the hyperparameters provided in Appendix~\ref{hyper}. The descriptive statistics of the main variables are reported in Table~\ref{tab:descstats}. The training samples cover 27 pipeline sections and the test samples cover 10, and the two sets have no section in common, so the test pipeline is unseen in every respect during training. All hyperparameters, including the number of layers, are selected on a validation subset held out from the 1781 training samples, and every model we compare is fitted and tuned under this protocol. The test labels are never used for fitting or for hyperparameter selection. They are read only for the final evaluation of Sections~\ref{comparison} and~\ref{ablation} and for the diagnostics that we identify as post hoc, namely Table~\ref{tab:icc}, the test columns of Table~\ref{tab:descstats}, Figure~\ref{fig:Probability Density} and Appendix~\ref{loso}, all of which were added in response to the review and none of which informed the model. The test features are read once more at inference, since Section~\ref{predict} predicts the test samples as one graph, and the protocol is transductive in that respect. Two of the $2309$ samples, one in the training set and one in the test set, fall below the threshold $10000e^{2.72\sqrt{d}}$ and are therefore evaluated on the second branch of Eqs.~(\ref{austin}) and~(\ref{Le}), while the remaining samples are evaluated on the first.

Two conventions govern the numerical treatment of the data. As to units, Table~\ref{tab:descstats} reports $d$ and $L_0$ in millimeters and kilometers as recorded, and both are converted to meters before Eqs.~(\ref{austin}) and~(\ref{Le}) are evaluated, whence $C_0$, $C_{\mathrm{AP}}$, $C_{\mathrm{AC}}$ and $C_{\Delta}$ are in meters throughout. As to scale, the columns of $\mathbf{X}$ span several orders of magnitude, $Re$ reaching $8.8\times10^{5}$ while $O$ is of order $10^{2}$, such that a message built from the recorded values would be governed by the Reynolds number alone were the columns left as recorded. Following \citet{you2020}, we therefore replace $\mathbf{X}_{\alpha\beta}$ by $(\mathbf{X}_{\alpha\beta}-x^{\min}_{\beta})/(x^{\max}_{\beta}-x^{\min}_{\beta})$ before the edge values of Section~\ref{sec:sdigrl} are formed, where $x^{\min}_{\beta}$ and $x^{\max}_{\beta}$ denote the extremes of column $\beta$ over the training samples alone. These constants are then applied unchanged to the validation and the test samples, whence the transformation is independent of the data on which the model is evaluated. Only the training values are therefore confined to $[0,1]$, and a validation or test value falls outside that interval whenever it exceeds the training range, as the minima of $d$, $L_0$ and $Re$ in Table~\ref{tab:descstats} do. The target column is transformed in the same way for the loss of Eq.~(\ref{loss}), and $\hat C_{\Delta,\alpha}$ of Eq.~(\ref{predict_delta}) is returned to meters by the inverse of that transformation before Eq.~(\ref{predict_ac}) is applied. All errors reported in Section~\ref{sec:case} are accordingly in meters.
The test pipeline is not a repetition of the training pipelines. Its inner diameters, transportation distances, and Reynolds numbers reach values the training set does not contain, and its mean $C_{\mathrm{AC}}$ is lower by about one third. The evaluation therefore measures how a model transfers to a pipeline it has never observed, which is the setting a newly commissioned pipeline presents in practice.
\begin{table}[pos=h]
\centering
\caption{Descriptive statistics of the variables, reported separately for the training set and the test set. The unit of $O$ is \(\mathrm{m}^3\), that of $d$ is \(\mathrm{mm}\), that of $L_0$ is \(\mathrm{km}\), and that of $C_{\mathrm{AP}}$, $C_{\mathrm{AC}}$ and $C_{\Delta}$ is \(\mathrm{m}\). The Reynolds number is dimensionless.}
\label{tab:descstats}
\begin{tabular}{lrrrrrrrr}
\toprule
& \multicolumn{4}{c}{Training set ($n_{\mathrm{train}}=1781$)} & \multicolumn{4}{c}{Test set ($n_{\mathrm{test}}=528$)} \\
\cmidrule(lr){2-5}\cmidrule(lr){6-9}
Variable & Min & Max & Mean & Std & Min & Max & Mean & Std \\
\midrule
$O$               & 0.0   & 651.0             & 410.6             & 136.4             & 26.0              & 219.0             & 96.6              & 36.5 \\
$d$               & 266.7 & 500.9             & 437.8             & 39.2              & 213.5             & 399.3             & 290.6             & 67.6 \\
$L_0$             & 63.6  & 228.2             & 112.1             & 43.9              & 32.1              & 116.8             & 52.5              & 19.8 \\
$Re$              & $1.3\times10^{5}$ & $8.8\times10^{5}$ & $4.5\times10^{5}$ & $1.5\times10^{5}$ & $1.8\times10^{4}$ & $5.0\times10^{5}$ & $2.8\times10^{5}$ & $8.8\times10^{4}$ \\
$C_{\mathrm{AP}}$ & 634.8 & $5.3\times10^{3}$ & $2.9\times10^{3}$ & 918.8             & 644.6             & $4.7\times10^{3}$ & $1.8\times10^{3}$ & 972.5 \\
$C_{\mathrm{AC}}$ & 745.6 & $5.5\times10^{3}$ & $3.0\times10^{3}$ & 900.0             & 807.0             & $5.1\times10^{3}$ & $2.1\times10^{3}$ & $1.1\times10^{3}$ \\
$C_{\Delta}$      & $-133.3$ & 907.4          & 125.7             & 116.8             & $-106.9$          & $1.0\times10^{3}$ & 243.4             & 193.9 \\
\bottomrule
\end{tabular}
\end{table}

\subsection{Evaluation Metrics}
\label{evaluation}
Let $C_{\mathrm{AC}}=(C_{\mathrm{AC},n_{\mathrm{train}}+1},\ldots,C_{\mathrm{AC},n})^{\top}$
and $\hat{C}_{\mathrm{AC}}=(\hat{C}_{\mathrm{AC},n_{\mathrm{train}}+1},\ldots,\hat{C}_{\mathrm{AC},n})^{\top}$
denote the vectors of true and predicted mixed oil lengths over the $n_{\mathrm{test}}$ test samples.
We employ three primary metrics to quantitatively evaluate the prediction accuracy. 
\begin{itemize}
    \item \textbf{Root Mean Square Error:} Root Mean Square Error (RMSE) measures the square root of the average squared difference between true values and predictions~\citep{hodson2022root}. The RMSE is:
\begin{equation}
\text{RMSE} = \frac{1}{\sqrt{n_{\text{test}}}} \left\| C_{\text{AC}} - \hat{C}_{\text{AC}} \right\|_2,
\label{RMSE}
\end{equation}
   where $\|\cdot\|_2$ denotes the $\ell_2$ norm. Lower RMSE values indicate better prediction accuracy.
    \item \textbf{Coefficient of Determination:} Coefficient of Determination (\(R^2\)) focuses on the proportion of variance explained by the model~\citep{ozer1985correlation}. Under the same notation as above, the \(R^2\) is:
    \begin{equation}
    R^2 = 1 - \frac{\left\| C_{\text{AC}} - \hat{C}_{\text{AC}} \right\|_2^2}{\left\| C_{\text{AC}} - \overline{C}_{\text{AC}} \right\|_2^2},
    \label{R^2}
\end{equation}
    where $\overline{C}_{\mathrm{AC}}=\frac{1}{n_{\mathrm{test}}}\sum_{\alpha=n_{\mathrm{train}}+1}^{n}C_{\mathrm{AC},\alpha}$. Higher \(R^2\) values indicate better prediction accuracy. Since the denominator of Eq.~(\ref{R^2}) depends on the test labels alone, it is common to every model, such that the values of $R^2$ reported in Tables~\ref{tab:1} to~\ref{tab:ablation} are all obtained from the same test vector and stand in a fixed relation to the corresponding RMSE.
    \item \textbf{Mean Absolute Error:} Mean Absolute Error (MAE) measures the average absolute difference between true values and predictions~\citep{hodson2022root}. Under the same notation as above, the MAE is:
\begin{equation}
\text{MAE} = \frac{1}{n_{\text{test}}} \left\| C_{\text{AC}} - \hat{C}_{\text{AC}} \right\|_1,
\label{MAE}
\end{equation}
where $\|\cdot\|_1$ denotes the $\ell_1$ norm. Lower MAE values indicate better prediction accuracy.

\end{itemize}

\begin{figure}[pos=h]
    \centering
    \includegraphics[width=0.8\textwidth]{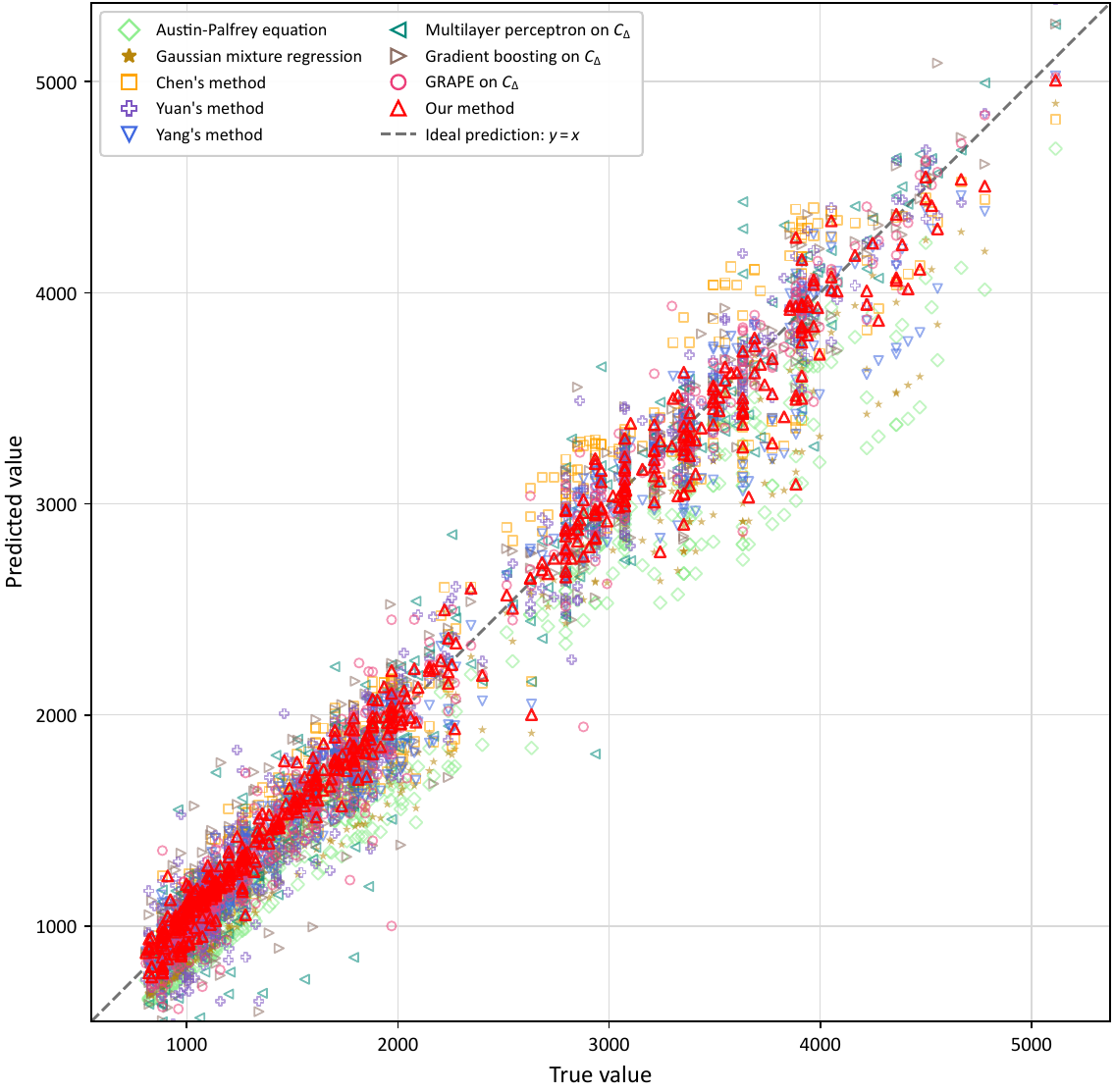}
    \caption{Comparison between predicted and true mixed oil lengths on the test set.}
    \label{fig:Predicted vs.}
\end{figure}

\subsection{Comparison with Baselines}
\label{comparison}
\begin{figure}[pos=h]
    \centering
    \includegraphics[width=0.95\textwidth]{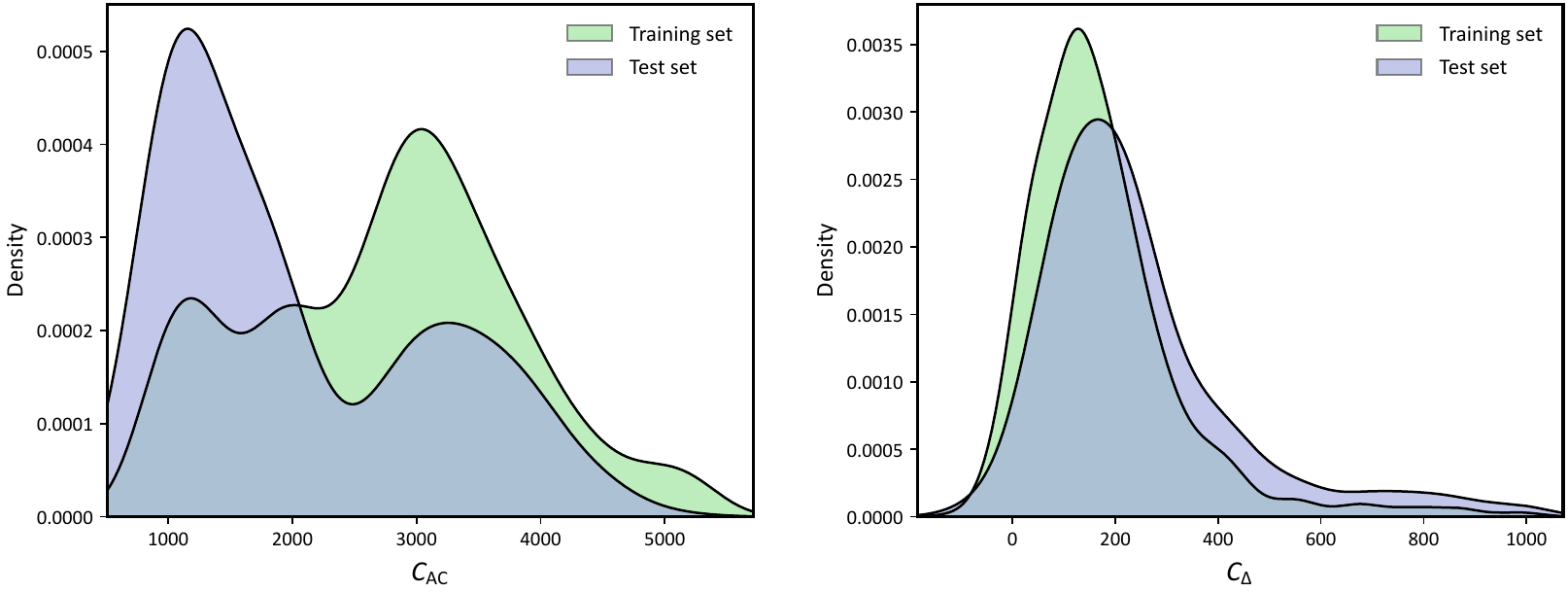}
    \caption{
         Left panel: estimated density functions of \(C_\mathrm{AC}\) in the training and test sets. Right panel: estimated density functions of $C_\Delta$ in the training and test sets.
    }
    \label{fig:Probability Density}
\end{figure}

We evaluate our method against five baselines, including the Austin-Palfrey equation, Gaussian mixture regression~\citep{yuan2023knowledge}, Chen's method~\citep{Chen2021b}, Yuan's SMR method~\citep{yuan2023}, and Yang's diffusion-based method~\citep{yang2024confidence}. We add three further baselines that isolate the contribution of the graph. A multilayer perceptron, a gradient boosting model~\citep{chen2016xgboost}, and the original GRAPE configuration of \citet{you2020} are all trained on the same residual target $C_\Delta$ with the same four features, so any gap between them and SDI-GRL is attributable to the graph structure and not to the residual formulation. The five earlier baselines are reproduced from the specifications given in the corresponding papers, and the three added ones use the reference implementations of their authors. Every baseline is tuned by grid search on the same validation subset and by the same criterion as SDI-GRL, the validation RMSE. The grid covers the depth and the width of the hidden layers together with the learning rate for the multilayer perceptron, the number of trees, the maximum depth, the learning rate and the subsampling ratio for gradient boosting, and the number of message passing layers, the embedding width and the learning rate for GRAPE, whose remaining settings are left at the defaults of \citet{you2020}. Appendix~\ref{hyper} reports the selected configuration of SDI-GRL. Figure~\ref{fig:Predicted vs.} provides a visual comparison of their prediction accuracy. The black line \(y=x\) represents perfect agreement between the predicted and true mixed oil lengths. The predictions produced by SDI-GRL are concentrated closest to this line, indicating the highest visual agreement with the true values. As shown in Table~\ref{tab:1}, SDI-GRL achieves the best performance on all three evaluation metrics. Compared with Yang's method, SDI-GRL reduces the RMSE from 171.913 to 138.563 and the MAE from 123.991 to 95.892, corresponding to relative reductions of 19.40\% and 22.66\%, respectively. It also increases \(R^2\) from 0.975 to 0.983. Among the three baselines trained to predict the same residual target \(C_\Delta\), GRAPE performs best, with an RMSE of 156.382, an MAE of 107.372, and an \(R^2\) of 0.979. Relative to this strongest target-matched baseline, SDI-GRL reduces the RMSE and MAE by 11.39\% and 10.69\%, respectively, while increasing \(R^2\) by 0.004. These results indicate that the performance gain cannot be attributed solely to the use of the residual prediction target and are consistent with the benefit of modeling dependencies among samples. Figure~\ref{fig:Probability Density} further shows that the training and test distributions of \(C_\Delta\) are more closely aligned than those of \(C_{\mathrm{AC}}\), suggesting that residual prediction reduces the marginal train-test distribution shift in the prediction target.

\begin{table}[pos=h]
\centering
\caption{Comparison of models with three evaluation metrics}
\label{tab:1}
\begin{tabular}{cccc}
\toprule
\textbf{Models} & \textbf{RMSE} & \boldmath$R^2$ & \textbf{MAE} \\
\midrule
Austin-Palfrey equation & 311.065 & 0.917 & 245.037 \\
Gaussian mixture regression~\citep{yuan2023knowledge} & 244.938 & 0.948 & 177.783 \\
Chen's method~\citep{Chen2021b} & 207.007 & 0.963 & 160.773 \\
Yuan's method~\citep{yuan2023} & 185.849 & 0.970 & 144.543 \\
Yang's method~\citep{yang2024confidence} & 171.913 & 0.975 & 123.991 \\
Multilayer perceptron on $C_\Delta$ & 198.326 & 0.966 & 140.278 \\
Gradient boosting on $C_\Delta$~\citep{chen2016xgboost} & 189.556 & 0.969 & 146.635 \\
GRAPE on $C_\Delta$~\citep{you2020} & 156.382 & 0.979 & 107.372 \\
\textbf{Our method} & \textbf{138.563} & \textbf{0.983} & \textbf{95.892} \\
\bottomrule
\end{tabular}
\end{table}

\subsection{Sensitivity Analysis}
\label{ablation}

We first examine sensitivity to the number of message-passing layers \(L\), which determines the number of information-exchange rounds among dependent samples. Table~\ref{tab:depth} reports the training and test performance for \(L\in\{2,4,6,8\}\), all other settings being fixed and \(L=6\) being the value selected on the validation subset of Section~\ref{sec_dataset}. Test performance improves appreciably as \(L\) increases from 2 to 4, the test RMSE falling by 2.18\%, whereas from \(L=4\) onward it is essentially flat, the three deeper settings differing by at most 0.667 in test RMSE and 0.717 in MAE, that is by less than 0.5\% and 0.8\% respectively, and \(R^2\) remaining between 0.983 and 0.984 throughout. The training RMSE, by contrast, decreases monotonically from 134.825 to 128.623, such that the train-test gap widens from 6.798 to 9.273. Since over-smoothing would manifest itself as a test error that rises with depth, and since no such rise is observed within the examined range, these results provide no evidence of pronounced over-smoothing, the widening gap being attributable instead to the additional capacity that greater depth confers. The three deeper settings are in any case too close to be distinguished on the test set, from which it follows that the performance reported below does not rest on a particular choice of \(L\).

\begin{table}[pos=htbp]
\centering
\caption{Performance under different numbers of message passing layers}
\label{tab:depth}
\begin{tabular}{ccccc}
\toprule
Layers \(L\) & Train RMSE & Test RMSE & \(R^2\) & MAE \\
\midrule
\(L=2\) & 134.825 & 141.623 & 0.983 & 97.735 \\
\(L=4\) & 131.278 & 138.542 & 0.983 & 95.672 \\
\(L=6\) & 129.994 & 138.563 & 0.983 & 95.892 \\
\(L=8\) & 128.623 & 137.896 & 0.984 & 95.175 \\
\bottomrule
\end{tabular}
\end{table}

We then analyze the sensitivity of SDI-GRL to three architectural and optimization choices, including the aggregation function, activation function, and optimizer. We compare Mean with Max aggregation, ReLU with Tanh activation, and Adam with RMSprop optimization~\citep{kingma2014adam,ruder2016overview}. Detailed descriptions of these components are provided in Appendix~\ref{keycomponents}.
Table~\ref{tab:ablation} reports the performance and running time of all eight settings. Setting 1 achieves the best overall performance with an RMSE of 138.563, an \(R^2\) of 0.983, and an MAE of 95.892, which confirms the Max, ReLU, and Adam configuration used in the main experiments. Setting 8 requires the shortest running time of 364.21 seconds, while Setting 4 requires the longest running time of 430.05 seconds. The results also indicate that performance depends on the interaction among aggregation, activation, and optimization choices. In particular, Adam performs better with ReLU, whereas RMSprop performs better with Tanh under both aggregation functions. Nevertheless, all eight settings outperform Yang's method in RMSE, \(R^2\), and MAE. Six of them, all but Settings 3 and 6, further attain a lower RMSE and a lower MAE than GRAPE, the strongest target-matched baseline, and match or exceed its \(R^2\) as well. Settings 3 and 6 each pair an activation function with the optimizer that suits it less well, in the sense just described, which accounts for their being the two weakest of the eight. These results demonstrate that SDI-GRL maintains strong predictive performance across the evaluated configurations.

\begin{table}[pos=htbp]
\centering
\caption{Performance under different sensitivity settings}
\label{tab:ablation}
\begin{tabular}{c|ccc|cccc}
\hline
\multirow{2}{*}{\textbf{Setting}}
& \multicolumn{3}{c|}{\textbf{Sensitivity Settings}}
& \multicolumn{4}{c}{\textbf{Performance}} \\
\cline{2-8}
& \textbf{Aggregation}
& \textbf{Activation}
& \textbf{Optimizer}
& \textbf{RMSE}
& \textbf{\(R^2\)}
& \textbf{MAE}
& \textbf{Time (s)} \\
\hline
1 & Max & ReLU & Adam & \textbf{138.563} & \textbf{0.983} & \textbf{95.892} & 409.92 \\

2 & Max & ReLU & RMSprop & 143.333 & 0.982 & 102.658 & 413.90 \\

3 & Max & Tanh & Adam & 160.928 & 0.978 & 110.059 & 421.56 \\

4 & Max & Tanh & RMSprop & 156.084 & 0.979 & 103.700 & 430.05 \\

5 & Mean & ReLU & Adam & 142.775 & 0.982 & 97.398 & 371.99 \\

6 & Mean & ReLU & RMSprop & 156.437 & 0.979 & 108.292 & 369.25 \\

7 & Mean & Tanh & Adam & 153.269 & 0.980 & 107.080 & 375.03 \\

8 & Mean & Tanh & RMSprop & 152.670 & 0.980 & 101.636 & \textbf{364.21} \\
\hline
\end{tabular}
\end{table}

\section{Conclusion}
\label{sec:conclu}
Accurate estimation of the length of mixed oil segments carries substantial
economic and operational implications for multi-product pipeline
operations~\citep{du2023deeppipe, sunagatullin2019relevant}. To address this
critical challenge, we propose the SDI-GRL model, a graph representation
learning approach that turns the statistical dependencies among
mixed oil samples into a usable learning signal. To the best of our knowledge, our work represents a
pioneering effort to systematically model and utilize the statistical
dependencies among mixed oil samples in predicting the mixed oil length. Specifically, through multi-layer message passing over the sample-feature graph, each prediction incorporates information from statistically dependent samples, a source of signal that prior methods discard by assuming independence. Quantitative experiments demonstrate that SDI-GRL reduces the RMSE and
MAE by 11.39\% and 10.69\% relative to the strongest existing baseline, and
the sensitivity analysis further shows that all eight evaluated configurations
outperform Yang's method across all three evaluation metrics.
It is worth emphasizing the practical implications of these gains.
More accurate length estimates permit operators to isolate the contaminated
segment with tighter cutting margins, thereby reducing the volume of
on-specification product that is unnecessarily downgraded and lowering the
reprocessing and storage costs associated with mixed oil handling. Moreover,
SDI-GRL draws solely on operational data that SCADA systems already collect,
allowing these gains to be attained without the installation and maintenance
costs of hardware-based detection instruments. Since the test samples originate
from a pipeline unseen during training, our results suggest that such a
transfer to a newly commissioned pipeline is possible. The evidence rests on one
held out pipeline evaluated as a single batch, however, so that deployment to a
pipeline whose records arrive a few at a time, or under an operating regime
unlike those observed here, remains to be examined. The graph constructed here encodes the statistical relation among samples rather than the physical layout of a pipeline, and an explicit encoding of that layout constitutes a natural direction for further development. Pipeline identity nodes, edges linking the samples of one section, and features describing the spatial and temporal state of the fluid could all be incorporated into the present construction, which would permit the transport process itself to be represented alongside the dependence among samples. Beyond mixed oil length estimation, SDI-GRL also paves the way for
promising extensions to other
practically essential and methodologically
compelling application domains, such as multi-phase flow
prediction~\citep{wen2019multiphase}.

\section*{Credit Authorship Contribution Statement}

\textbf{Xinyi Hong:} Writing, Original Draft, Visualization, Software, Methodology, Investigation, Conceptualization. \textbf{Hai Shu:} Review \& Editing, Methodology. \textbf{Zhaoyang Zhang:} Writing, Review \& Editing, Software. \textbf{Lei Chen:} Data curation. \textbf{Yanfeng Yang*:} Writing, Review \& Editing, Visualization, Methodology. \textbf{Ziqi Chen*:} Writing, Review \& Editing, Methodology, Supervision.

\section*{Declaration of Competing Interest}

The authors declare that they have no known competing financial interests or personal relationships that could have appeared to influence the work reported in this paper.

\section*{Data Availability}

The data that has been used is confidential. 
\section*{Acknowledgments}

Dr. Ziqi Chen's work was partially supported by the National Key R\&D Program of China (2021YFA1000100 and 2021YFA1000101) and National Natural Science Foundation of China (NSFC) (12271167). Yanfeng Yang's work was supported by JST SPRING of Japan (JPMJSP2104).

\bibliography{references}

@article{yuan2023,
  title={Physics-informed Student’st mixture regression model applied to predict mixed oil length},
  author={Yuan, Ziyun and Chen, Lei and Liu, Gang and Shao, Weiming and Zhang, Yuhan and Ma, Yunxiu},
  journal={Journal of Pipeline Science and Engineering},
  volume={3},
  number={1},
  pages={100105},
  year={2023},
  publisher={Elsevier}
}

@article{you2020,
  title={Handling missing data with graph representation learning},
  author={You, Jiaxuan and Ma, Xiaobai and Ding, Yi and Kochenderfer, Mykel J and Leskovec, Jure},
  journal={Advances in Neural Information Processing Systems},
  volume={33},
  pages={19075--19087},
  year={2020}
}

@inproceedings{baptista2001,
  title={Mixing volume evaluation under pumping shut-down conditions},
  author={Baptista, Renan Martins and Rachid, Felipe Bastos de Freitas and de Araujo, Jose Henrique Carneiro},
  booktitle={PSIG Annual Meeting},
  pages={PSIG--0111},
  year={2001},
  organization={PSIG}
}

@article{reniers2021,
  title={Safety and security of oil and gas pipeline transportation: A systematic analysis of research trends and future needs using WoS},
  author={Chen, Chao and Li, Changjun and Reniers, Genserik and Yang, Fuqiang},
  journal={Journal of Cleaner Production},
  volume={279},
  pages={123583},
  year={2021},
  publisher={Elsevier}
}

@article{Trench2001,
  title={How pipelines make the oil market work--Their networks, operation and regulation},
  author={Trench, Cheryl J},
  journal={Allegro Energy Group, New York},
  year={2001}
}

@article{Huang2021,
  title={Carbon footprint of oil products pipeline transportation},
  author={Huang, Liqiao and Liao, Qi and Yan, Jinyue and Liang, Yongtu and Zhang, Haoran},
  journal={Science of the Total Environment},
  volume={783},
  pages={146906},
  year={2021},
  publisher={Elsevier}
}

@article{Mirhassani2011,
  title={Scheduling multi-product tree-structure pipelines},
  author={MirHassani, SA and Jahromi, H Fani},
  journal={Computers \& chemical engineering},
  volume={35},
  number={1},
  pages={165--176},
  year={2011},
  publisher={Elsevier}
}

@article{Austin1963,
  title={Mixing of miscible but dissimilar liquids in serial flow in a pipeline},
  author={Austin, J and Palfrey, J},
  journal={Proceedings of the Institution of Mechanical Engineers},
  volume={178},
  number={1},
  pages={377--389},
  year={1963},
  publisher={SAGE Publications}
}

@article{Du2023a,
  title={Deeppipe: An intelligent framework for predicting mixed oil concentration in multi-product pipeline},
  author={Du, Jian and Zheng, Jianqin and Liang, Yongtu and Xia, Yuheng and Wang, Bohong and Shao, Qi and Liao, Qi and Tu, Renfu and Xu, Bin and Xu, Ning},
  journal={Energy},
  volume={282},
  pages={128810},
  year={2023},
  publisher={Elsevier}
}

@article{Chen2021b,
  title={A novel predictive model of mixed oil length of products pipeline driven by traditional model and data},
  author={Chen, Lei and Yuan, Ziyun and Xu, JianXin and Gao, Jingyang and Zhang, Yuhan and Liu, Gang},
  journal={Journal of Petroleum Science and Engineering},
  volume={205},
  pages={108787},
  year={2021},
  publisher={Elsevier}
}

@article{Patrachari2012,
  title={A conceptual framework to model interfacial contamination in multiproduct petroleum pipelines},
  author={Patrachari, Anirudh R and Johannes, Arland H},
  journal={International journal of heat and mass transfer},
  volume={55},
  number={17-18},
  pages={4613--4620},
  year={2012},
  publisher={Elsevier}}

@article{Li2023,
  title={An Industrial Scale Simulation Method for Predicting the Contamination Evolution in Long-Distance Multiproduct Pipelines},
  author={Li, Yansong and Han, Dong and Wang, Baoying and Sun, Jianfei},
  journal={Journal of Pipeline Systems Engineering and Practice},
  volume={14},
  number={1},
  pages={04022054},
  year={2023},
  publisher={American Society of Civil Engineers}
}

@article{Wang2020,
  title={Study on asymmetry concentration of mixed oil in products pipeline},
  author={Wang, Yi and Wang, Baoying and Liu, Yang and Liang, Yongtu},
  journal={Energies},
  volume={13},
  number={23},
  pages={6398},
  year={2020},
  publisher={MDPI}
}

@article{Hamidi2023,
  title={Numerical Simulation of Mixing between Sequential Fluids in a Pipeline Considering Real Conditions},
  author={Hamidi, Vali and Hajidavalloo, Ebrahim and Behbehani-Nejad, Morteza},
  journal={Chemical Engineering \& Technology},
  volume={46},
  number={4},
  pages={644--653},
  year={2023},
  publisher={Wiley Online Library}
}

@inproceedings{Li2021,
  title={Data-driven predictive model for mixed oil length prediction in long-distance transportation pipeline},
  author={Li, Xu and Han, Wenxue and Shao, Weiming and Chen, Lei and Zhao, Dongya},
  booktitle={2021 IEEE 10th Data Driven Control and Learning Systems Conference (DDCLS)},
  pages={1486--1491},
  year={2021},
  organization={IEEE}
}

@article{Friedman2001,
  title={Greedy function approximation: a gradient boosting machine},
  author={Friedman, Jerome H},
  journal={Annals of statistics},
  pages={1189--1232},
  year={2001},
  publisher={JSTOR}
}

@article{Yuan2021,
  title={A robust hybrid predictive model of mixed oil length with deep integration of mechanism and data},
  author={Yuan, Ziyun and Chen, Lei and Shao, Weiming and Zuo, Zhiheng and Zhang, Wan and Liu, Gang},
  journal={Journal of Pipeline Science and Engineering},
  volume={1},
  number={4},
  pages={459--467},
  year={2021},
  publisher={Elsevier}
}

@article{kipf2016semi,
  title={Semi-supervised classification with graph convolutional networks},
  author={Kipf, Thomas N and Welling, Max},
  journal={arXiv preprint arXiv:1609.02907},
  year={2016}
}

@article{yang2024confidence,
  title={Confidence interval estimation of mixed oil length with conditional diffusion model},
  author={Yang, Yanfeng and Zhang, Lihong and Chen, Ziqi and Yu, Miaomiao and Chen, Lei},
  journal={Journal of Pipeline Science and Engineering},
  volume={5},
  number={2},
  pages={100248},
  year={2025},
  doi={10.1016/j.jpse.2024.100248},
  publisher={Elsevier}
}

@article{du2023deeppipe,
  title={Deeppipe: A two-stage physics-informed neural network for predicting mixed oil concentration distribution},
  author={Du, Jian and Zheng, Jianqin and Liang, Yongtu and Xu, Ning and Kleme{\v{s}}, Ji{\v{r}}{\'\i} Jarom{\'\i}r and Wang, Bohong and Liao, Qi and Varbanov, Petar Sabev and Shahzad, Khurram and Ali, Arshid Mahmood},
  journal={Energy},
  volume={276},
  pages={127452},
  year={2023},
  publisher={Elsevier}
}

@article{yuan2023knowledge,
  title={Knowledge-informed Variational Bayesian Gaussian mixture regression model for predicting mixed oil length},
  author={Yuan, Ziyun and Chen, Lei and Liu, Gang and Zhang, Yuhan},
  journal={Energy},
  volume={285},
  pages={129248},
  year={2023},
  publisher={Elsevier}
}

@inproceedings{chen2016xgboost,
  title={Xgboost: A scalable tree boosting system},
  author={Chen, Tianqi and Guestrin, Carlos},
  booktitle={Proceedings of the 22nd acm sigkdd international conference on knowledge discovery and data mining},
  pages={785--794},
  year={2016}
}

@article{wen2019multiphase,
  title={Multiphase flow prediction with deep neural networks},
  author={Wen, Gege and Tang, Meng and Benson, Sally M},
  journal={arXiv preprint arXiv:1910.09657},
  year={2019}
}

@article{rumelhart1986learning,
  title={Learning representations by back-propagating errors},
  author={Rumelhart, David E and Hinton, Geoffrey E and Williams, Ronald J},
  journal={nature},
  volume={323},
  number={6088},
  pages={533--536},
  year={1986},
  publisher={Nature Publishing Group UK London}
}

@inproceedings{kingma2014adam,
  title={Adam: A method for stochastic optimization},
  author={Kingma, Diederik P and Ba, Jimmy},
  booktitle={3rd International Conference on Learning Representations (ICLR)},
  year={2015}
}

@article{sunagatullin2019relevant,
  title={Relevant issues on quality evaluation of petroleum pipeline preparation for oil product transportation},
  author={Sunagatullin, Rustam and Timofeev, Fedor and Kuznetsov, Andrey and Oludina, Yulia},
  journal={OIL \& GAS Science and Technology--Revue D’ifp Energies Nouvelles},
  volume={74},
  pages={44},
  year={2019},
  publisher={EDP Sciences}
}

@article{du2025deeppipe,
  title={Deeppipe: A physics-enhanced adaptive multi-modal fused neural network for predicting contamination length interval in multi-product pipeline},
  author={Du, Jian and Zheng, Jianqin and Niu, Pengtao and Pan, Shiyuan and Liao, Qi and Qiu, Rui and Xu, Bin and Xu, Ning and Liang, Yongtu},
  journal={Engineering Applications of Artificial Intelligence},
  volume={158},
  pages={111564},
  year={2025},
  publisher={Elsevier}
}

@inproceedings{he2015delving,
  title={Delving deep into rectifiers: Surpassing human-level performance on imagenet classification},
  author={He, Kaiming and Zhang, Xiangyu and Ren, Shaoqing and Sun, Jian},
  booktitle={Proceedings of the IEEE international conference on computer vision},
  pages={1026--1034},
  year={2015}
}

@inproceedings{hamilton2017inductive,
  title={Inductive Representation Learning on Large Graphs},
  author={Hamilton, William L. and Ying, Rex and Leskovec, Jure},
  booktitle={Advances in Neural Information Processing Systems (NeurIPS)},
  volume={30},
  year={2017}
}

@article{moreno2012unifying,
  title={A unifying view on dataset shift in classification},
  author={Moreno-Torres, Jose G and Raeder, Troy and Alaiz-Rodr{\'\i}guez, Roc{\'\i}o and Chawla, Nitesh V and Herrera, Francisco},
  journal={Pattern recognition},
  volume={45},
  number={1},
  pages={521--530},
  year={2012},
  publisher={Elsevier}
}

@article{ruder2016overview,
  title={An overview of gradient descent optimization algorithms},
  author={Ruder, Sebastian},
  journal={arXiv preprint arXiv:1609.04747},
  year={2016}
}

@article{hodson2022root,
  title={Root mean square error (RMSE) or mean absolute error (MAE): When to use them or not},
  author={Hodson, Timothy O},
  journal={Geoscientific Model Development},
  volume={15},
  number={14},
  pages={5481--5487},
  year={2022},
  doi={10.5194/gmd-15-5481-2022},
  publisher={Copernicus Publications}
}

@article{ozer1985correlation,
  title={Correlation and the coefficient of determination.},
  author={Ozer, Daniel J},
  journal={Psychological bulletin},
  volume={97},
  number={2},
  pages={307},
  year={1985},
  publisher={American Psychological Association}
}

@inproceedings{grover2016node2vec,
  title={node2vec: Scalable feature learning for networks},
  author={Grover, Aditya and Leskovec, Jure},
  booktitle={Proceedings of the 22nd ACM SIGKDD international conference on Knowledge discovery and data mining},
  pages={855--864},
  year={2016}
}

@article{abu2018watch,
  title={Watch your step: Learning node embeddings via graph attention},
  author={Abu-El-Haija, Sami and Perozzi, Bryan and Al-Rfou, Rami and Alemi, Alexander A},
  journal={Advances in neural information processing systems},
  volume={31},
  year={2018}
}

@article{srivastava2014dropout,
  title={Dropout: a simple way to prevent neural networks from overfitting},
  author={Srivastava, Nitish and Hinton, Geoffrey and Krizhevsky, Alex and Sutskever, Ilya and Salakhutdinov, Ruslan},
  journal={The journal of machine learning research},
  volume={15},
  number={1},
  pages={1929--1958},
  year={2014},
  publisher={JMLR. org}
}

@book{boyer2004scada,
  title={SCADA: supervisory control and data acquisition},
  author={Boyer, Stuart A},
  year={2004},
  publisher={John Wiley \& Sons}
}
\newpage
\appendix
\section*{Appendix}
\renewcommand{\thesection}{A\arabic{section}} 

\section{Empirical Analysis of the Statistical Dependence}
\label{loso}
This appendix specifies the two analyses summarized in Section~\ref{sec:intr}. Both use the pipeline section as the grouping unit, a section being a distinct pair of inner diameter and transportation distance, of which the dataset contains 37. The group sizes range from 2 to 176 samples, with a mean of $2309/37=62.4$.
The coefficient reported in Table~\ref{tab:icc} is the single measurement intraclass correlation of a one way random effects model. Writing $y_{g\alpha}$ for the $\alpha$-th sample of section $g$, the model is $y_{g\alpha}=\mu+u_g+\epsilon_{g\alpha}$ with $u_g$ and $\epsilon_{g\alpha}$ independent and of variances $\sigma^2_u$ and $\sigma^2_\epsilon$, and the coefficient is $\mathrm{ICC}=\sigma^2_u/(\sigma^2_u+\sigma^2_\epsilon)$. We estimate the two variance components by the unbalanced analysis of variance estimator, which compares the between section and the within section mean squares at the average group size, and we obtain the confidence interval and the $p$-value from the exact $F$ distribution of their ratio, the null hypothesis being $\sigma^2_u=0$. The third row of the table applies the same procedure to the residual of $C_\Delta$ from a least squares fit of $C_\Delta$ on $O$, $d$, $L_0$ and $Re$ with an intercept, the within section degrees of freedom being reduced by the five estimated coefficients, namely the four slopes and the intercept.
Two qualifications should be attached to this quantity. Since the adjustment is linear, a nonlinear conditional mean or an omitted interaction would leave structure in the residual that clusters by section, whence the figure is best read as the clustering that a linear adjustment fails to absorb rather than as conditional dependence in a strict sense. Moreover, a section is itself defined by a pair of values of $d$ and $L_0$, both of which are inputs, so that section membership is not independent of the features being adjusted for. Nevertheless, the magnitude of the residual coefficient is large enough to indicate that the samples of one section are not exchangeable with those of another once the recorded features are taken into account.

The ICC does not by itself show that this clustering carries information beyond the recorded features. To examine that question, we hold out one section at a time, fit a gradient boosting model on $O$, $d$, $L_0$ and $Re$ over the remaining sections, and apply the fitted model to the held out section. The model is a histogram based gradient boosting regressor left at its default settings and run from a fixed random seed, so that the diagnostic depends on no tuning choice of ours. If the effect of a section were a smooth function of these four features, the model would recover that effect from the other sections, and the residuals $r_\alpha$ on the held out section, which contains $n_g$ samples, would then be centered at zero. The observed residuals instead carry a systematic offset. Averaging the ratio $(\sum_\alpha r_\alpha/n_g)^2/(\sum_\alpha r_\alpha^2/n_g)$ over the held out sections with equal weight gives $32.9\%$. The quantity is thus a mean over sections of a per section ratio, and not the share of a pooled mean squared error. The same behavior appears on the split used in Section~\ref{sec_dataset}, where a model fitted on the 1781 training samples leaves an offset of $-68.59\,\mathrm{m}$ on the 528 test samples. We stress that such an offset is consistent with a section level effect that the recorded features do not transfer, but that it is equally consistent with model bias or with extrapolation beyond the range the fitted model has seen. The analysis is therefore a diagnostic of association, and it motivates conditioning a prediction on the cohort rather than establishing that SDI-GRL recovers an unobserved cause.

\section{The Training Graph and the Test Graph}
\label{flowchart}
\begin{figure}[pos=H]
    \centering
    \includegraphics[width=\textwidth]{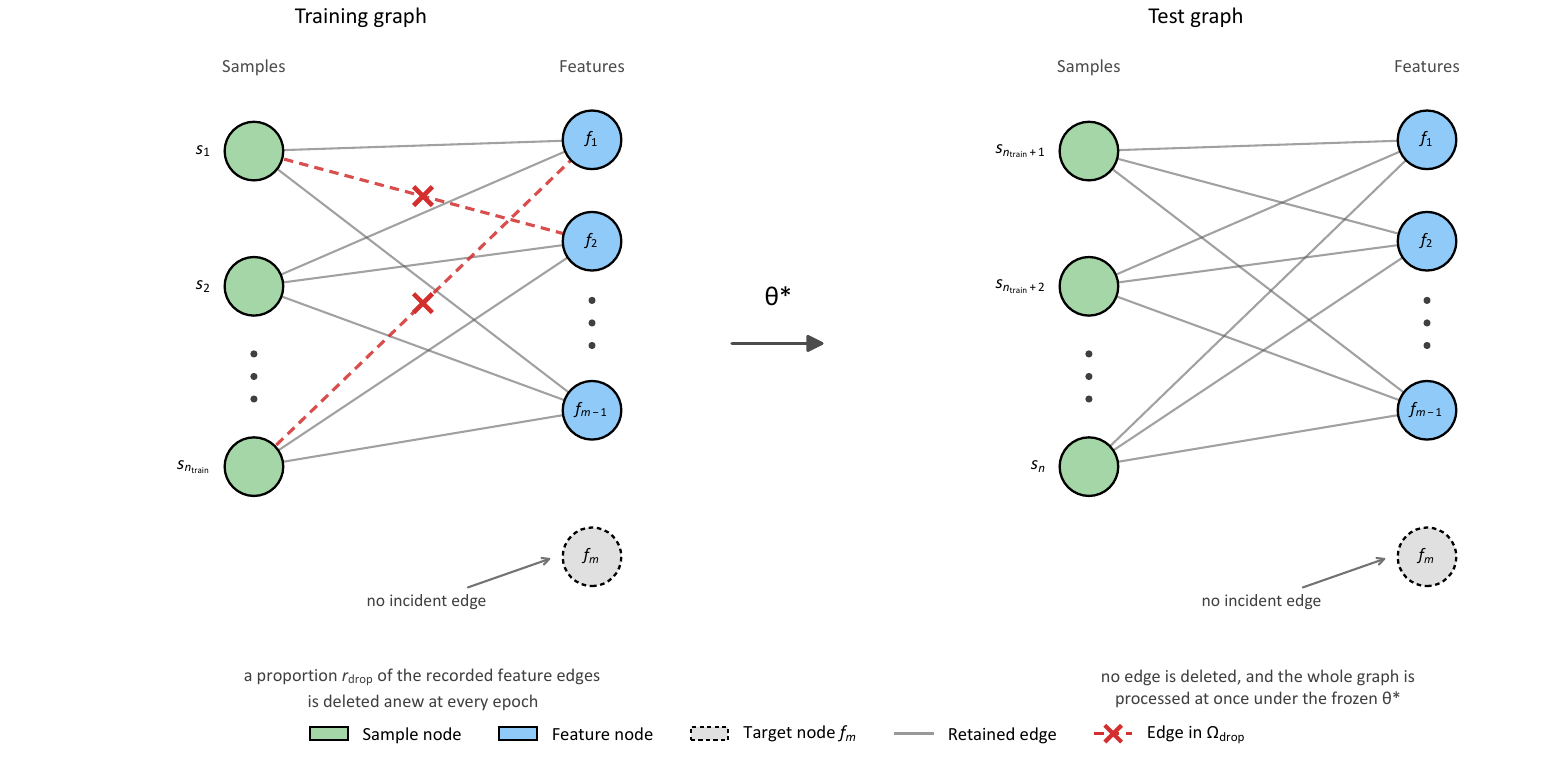}
    \caption{\textbf{The training graph and the test graph.} The training graph $\mathcal{G}_{\mathrm{train}}$ is shown on the left and the test graph $\mathcal{G}_{\mathrm{test}}$ on the right. The two are built from disjoint sets of samples and share the feature nodes of $\mathcal{V}_F$. The target node $f_m$ carries no incident edge in either graph, such that the target of a sample is absent from the input by construction rather than replaced by a substitute value, and the model therefore receives the same kind of input at inference as it does during training. On the training graph a proportion $r_{\mathrm{drop}}$ of the recorded feature edges is deleted anew at every epoch, whereas the test graph retains all of its edges and is processed as a whole under the optimized parameters $\theta^{*}$.}
    \label{fig:traintest}
\end{figure}

Section~\ref{sec:sdigrl} defines the training graph and the test graph algebraically. Figure~\ref{fig:traintest} places them side by side, from which the property they share and the two on which they differ can be read at once. They share the absence of any edge incident on the target node $f_m$, and they differ in the samples they are built from and in the edge dropout, which acts on the training graph alone.

\section{Implementation Details}
\label{hyper}
We report the detailed hyperparameter settings in Table~\ref{tab:2}. Two entries of the table follow from Eq.~(\ref{concat_layers}). The node representations $\tilde h_{s_\alpha}$ and $\tilde h_{f_\beta}$ concatenate the embeddings of the $L$ layers and are therefore of width $d_VL=96$, and the readout $\mathcal{H}$ of Eq.~(\ref{output}) receives the concatenation of two of them and is therefore of input width $2d_VL=192$. The parameter counts follow from these widths together with Eqs.~(\ref{msx}) to~(\ref{esxl}). The first message passing layer maps the node initializations of width $m$ and the scalar edge value, and therefore contributes $1{,}472$ parameters, whereas each of the five subsequent layers maps embeddings of width $d_V$ and $d_E$ and contributes $2{,}896$, which gives $15{,}952$ in total. Adding the $33{,}248$ parameters of $\mathcal{P}$ and the $33{,}217$ of $\mathcal{H}$ yields $82{,}417$, so that the reported capacity can be reconstructed from the description of the model.

\section{Computational Hardware}
\label{hard}
The models are trained on a Linux 64-bit OS with about 660 GiB of RAM, an Intel Xeon Platinum 8470Q CPU with 208 logical cores, and 1 NVIDIA GeForce RTX 5090 GPU with 31.84 GiB of VRAM.

\begin{table}[pos=h]
    \centering
    \caption{Hyperparameter Settings}
    \label{tab:2}
    \begin{tabular}{ll}
        \toprule
        \textbf{Category} & \textbf{Hyperparameter Setting} \\
        \midrule
        \textbf{Network Structure} & \\
        \quad Number of layers \(L\) & 6 \\
        \quad Embedding dimensions \(d_V\), \(d_E\), \(d_M\) & 16, 16, 16 \\
        \quad MLPNet \(\mathcal{P}\), input, hidden and output widths & \(d_VL=96\), six layers of 64, \(d_VL=96\) \\
        \quad Readout MLPNet \(\mathcal{H}\), input, hidden and output widths & \(2d_VL=192\), six layers of 64, 1 \\
        \quad Trainable parameters, message passing layers & 15{,}952 \\
        \quad Trainable parameters, \(\mathcal{P}\) and \(\mathcal{H}\) & 33{,}248 and 33{,}217 \\
        \quad Total number of trainable parameters & 82{,}417 \\
        \quad Weight initialization (\(W_1\) to \(W_5\)) & Kaiming's method~\citep{he2015delving} \\
        \midrule
        \textbf{Model Components} & \\
        \quad Aggregation function & Max \\
        \quad Activation function & ReLU \\
        \midrule
\textbf{Training Configuration} & \\
\quad Learning rate & 0.001 \\
\quad Optimizer & Adam \\
\quad Number of training epochs & 3000 \\
\quad Edge dropout ratio \(r_{\mathrm{drop}}\) & \(\text{0.20}\) \\
\quad Learning rate decay step & 1000 \\
\quad Learning rate decay rate & 0.9 \\
        \bottomrule
    \end{tabular}
\end{table}

\section{Key Components in Sensitivity Analysis}
\label{keycomponents}

We provide detailed descriptions of three components considered in the sensitivity analysis: aggregation functions, activation functions, and optimizers.

\begin{enumerate}
  \item \textbf{Aggregation functions.} 
  Let $X \in \mathbb{R}^{N \times M}$ be a matrix with rows $\{x_1, \dots, x_N\}$. 
  The mean aggregation is defined as
  \[
  \mathrm{MEAN}(X) = \frac{1}{N} \sum_{i=1}^{N} x_i,
  \]
  and the max aggregation is defined as
  \[
  \mathrm{MAX}(X) = \max_{1 \le i \le N} x_i,
  \]
  where the maximum is taken element-wise across rows.

  \item \textbf{Activation functions.} 
  For a matrix $X = (x_{ij}) \in \mathbb{R}^{N \times M}$, activation functions are applied element-wise. 
  The ReLU function is
  \[
  \mathrm{ReLU}(X) = (\max(0, x_{ij})) \in \mathbb{R}^{N \times M},
  \]
  and the hyperbolic tangent function is
  \[
  \tanh(X) = \left(\frac{e^{x_{ij}} - e^{-x_{ij}}}{e^{x_{ij}} + e^{-x_{ij}}}\right),
  \]
  which maps each entry to the range $(-1, 1)$.

  \item \textbf{Optimizers.} 
  The optimizer updates model parameters during training. 
  Adam is an adaptive method based on estimates of the first and second moments of the gradients, providing parameter-wise learning rate adjustments. 
  RMSprop scales the learning rate by a moving average of recent squared gradients, improving stability and convergence efficiency.
\end{enumerate}

\end{document}